\documentclass[sigconf]{aamas}

\usepackage{balance} 
\usepackage{natbib}                 
\usepackage{subcaption}             
\usepackage{amsmath}                
\usepackage{multirow}
\usepackage{multicol}
\usepackage{array}
\usepackage{capt-of}
\usepackage{algorithm}
\usepackage{algorithmic}
\usepackage{placeins}               

\makeatletter
\gdef\@copyrightpermission{
  \begin{minipage}{0.2\columnwidth}
    \href{https://creativecommons.org/licenses/by/4.0/}{\includegraphics[width=0.90\textwidth]{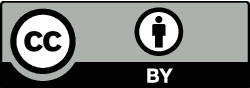}}
  \end{minipage}\hfill
  \begin{minipage}{0.8\columnwidth}
    \href{https://creativecommons.org/licenses/by/4.0/}{This work is licensed under a Creative Commons Attribution International 4.0 License.}
  \end{minipage}
  \vspace{5pt}
}
\makeatother

\title[AAMAS-2025 Formatting Instructions]{Distributed Value Decomposition Networks with Networked Agents}

\author{Guilherme S. Varela}
\affiliation{
 \institution{Instituto Superior Técnico, INESC-ID}
 \city{Lisbon}
 \country{Portugal}}
\email{guilherme.varela@tecnico.ulisboa.pt}

\author{Alberto Sardinha}
\affiliation{
 \institution{PUC-Rio}
 \city{Rio de Janeiro}
 \country{Brazil}}
\email{sardinha@inf.puc-rio.br}

\author{Francisco S. Melo}
\affiliation{
 \institution{Instituto Superior Técnico, INESC-ID}
 \city{Lisbon}
 \country{Portugal}}
\email{fmelo@inesc-id.pt}

\begin{abstract}
We investigate the problem of distributed training under partial observability, whereby cooperative multi-agent reinforcement learning agents (MARL) maximize the expected cumulative joint reward. We propose distributed value decomposition networks (DVDN) that generate a joint Q-function that factorizes into agent-wise Q-functions. Whereas the original value decomposition networks rely on centralized training, our approach is suitable for domains where centralized training is not possible and agents must learn by interacting with the physical environment in a decentralized manner while communicating with their peers. DVDN overcomes the need for centralized training by locally  estimating the shared objective. We contribute with two innovative algorithms, DVDN and DVDN (GT), for the heterogeneous and homogeneous agents settings respectively. Empirically, both algorithms approximate the performance of value decomposition networks, in spite of the information loss during communication, as demonstrated in ten MARL tasks in three standard environments.
\end{abstract}

\keywords{Artificial Intelligence, Machine Learning, Multi-Agent Systems, Reinforcement Learning, Deep Learning}

\newcommand{\TDR}[1]{\mathbb{E}^{\pi}_{\substack{s^0\sim\mu(\cdot)}}\left[#1\right]}

\newcommand{\loss}[1]{\ell\big(#1\big)} 
\newcommand{\RR}{\mathbb{R}} 

\newcommand{\ie}{\emph{i.e.}}
\newcommand{\eg}{\emph{e.g.}}

\renewcommand{\S}{\mathcal{S}}
\newcommand{\A}{\mathcal{A}}

\newcommand{\E}{\mathcal{E}}
\newcommand{\G}{\mathcal{G}}
\renewcommand{\H}{\mathcal{H}}
\newcommand{\N}{\mathcal{N}}
\newcommand{\M}{\mathcal{M}}

\renewcommand{\P}{\mathcal{P}}
\renewcommand{\O}{\mathcal{O}}

\newcommand{\JN}{j\in\N^{(k)}_i}   

\newcommand{\CI}[2]{$\begin{array}{c}{#1}\\({#2})\end{array}$}  
\newcommand{\CB}[2]{$\begin{array}{c}\mathbf{{#1}}\\({#2})\end{array}$} 

\newtheorem{fact}{Fact}

\begin{document}


\pagestyle{fancy}
\fancyhead{}


\maketitle 


\section{Introduction}

Cooperative multi-agent reinforcement learning addresses the problem of designing utility-maximizing agents that learn by interacting with a shared environment. Representing utility functions and applying them for decision-making is challenging because of the large combined joint observation and joint action spaces. Value decomposition network (VDN)~\citep{sunehag_2018} avoid this combinatorial trap by considering the family of $Q$-functions that factorize agent-wise. The method offers a viable solution to the scalability of MARL systems under the premise of {\em centralized training with decentralized execution}, where the outputs of the individual $Q$-functions are added to form a joint $Q$-function.

However, in many real-world domains, the premise of centralized training is too restrictive. For instance, in reinforcement learning based distributed\footnote{We use distributed and decentralized interchangeably.} load balancing~\citep{zheng_2023} intelligent switches act as agents to distribute multiple types of requests to a fleet of servers in a data center. Agents assign the incoming load to the servers, resolving requests at low latencies under quality-of-service constraints. In this domain, there is no simulator---agents must learn online by observing queues at links and selecting the links to route the requests. In robotic teams, there might be only simulators with a considerable gap between the simulated environment and the real-world. In real-world situations where communication is restricted and actions can fail in unpredictable ways, also benefit from this approach. In {\em RoboCup}~\citep{antonioni_2021} tournament, soccer robots actuate as agents and are endowed with sensors and computation onboard. Although communication has delays and link failures, agents should cooperate as a team to score goals.

The straightforward alternative to centralized training is the {\em fully decentralized training} approach, employing {\em independent $Q$-learners} (IQL)~\citep{tan_1993}. As IQL agents are oblivious to the presence of their teammates, they cannot account for the joint action, and from the perspective of any single agent the perceived transition probabilities of the environment are non-stationary. This approach violates the reinforcement learning assumption that the transition probabilities are fixed and unknown. Since individual learners do not communicate, fully decentralized IQL precludes {\em parameter sharing}, where agents update the same policy parameters. For many MARL tasks, parameter sharing improves sample efficiency but requires that every agent updates the same weights. Hence in a {\em decentralized training} setting, every agent would need one-to-one communication to broadcast its weights and experiences to all other agents, before performing the updates locally.

We propose a novel algorithm that combines the decentralized training with value decomposition networks' $Q$-function decomposition. Starting from the loss function used in VDN, in centralized training setting, we show that information back-propagated to agents' neutral network is the {\em joint temporal difference} (JTD). Our algorithm operates in the {\em decentralized training and decentralized execution} (DTDE) setting where agents communicate with their closest neighbors to improve their local JTD estimations. And each agent locally minimizes JTD. When agents are homogeneous, \ie, having the same individual observation and action space, we incentivize them to {\em align} their $Q$-function's parameters and their gradients; This mechanism called {\em gradient tracking}~\citep{qu_2018} enables agents to minimize a common loss function. To the best of our knowledge, this is the first application of gradient tracking for policy iteration in reinforcement learning.

\section{Background}

We introduce three concepts from the {\em distributed optimization} literature~\citep{nedic_2020}  that form the key components of our method: (i) The {\em switching topology communication channel} over which agents perform communication with closest neighbors. (ii) The consensus updates over the switching topology communication channel, where agents agree on the value of a constant; (iii) and finally, gradient tracking that allows agents to optimize a global function formed by the sum of local functions, using local computation and communication. Additionally, we define {\em decentralized partially observable Markov decision process} the mathematical framework underpinning value decomposition network.

\textbf{Switching topology communication channel}: Given a set $\N=\{1,\cdots, N\}$ of agents connected in a communication network such that agents $i$ and $j$ can exchange messages if and only if $i, j\in \E$ where $\E\subseteq \N\times \N$ denotes the edge set. Under the switching topology regime at every round of communication $k$, there is a different set of edges $\E^{(k)}$, such that agents are uncertain in advance who their peers are going to be. This randomness is captured by a underlying {\em time-varying undirected graph} $\G^{(k)}=\G^{(k)}(\N, \E^{(k)})$ where the node set $\N$ remains fixed but the edge set $\E^{(k)}$ is allowed to vary at each timestep $k$.

\textbf{Consensus over undirected time-varying networks~\citep{xiao_2007}}: is the process by which nodes initially holding scalar values asymptotically agree on their network average by interacting with closest neighbors over a switching network topology. At each iteration, each node replaces its own value with the weighted average of its previous value and the values of its neighbors. More formally, let the variable $x_i^{(0)}$ be held by agent $i$. Then, agents agree on a value $\bar{x}$ by performing the updates: 
\begin{equation}\label{eqn:consensus} x^{(k+1)} = \sum_{j\in\N^{(k)}_i} \alpha^{(k)}_{i, j}x^{(k)}_j\text{,}\end{equation}
where $\N^{(k)}_i = \left\{i\cup j| (i, j)\in \E^{(k)}\right\}$ is the {\em neighborhood of agent i} at communication step $k$, extended to include $x_i$ measurement as well. Under the time-varying regime we allow the neighborhood to switch at every consensus update. A foundational result~~\citep{xiao_2007} establishes the existence of the weights $\left[\alpha^{(k)}_{i, j}\right]_{N\times N}$, such that by repeating the updates in Eqn.~\eqref{eqn:consensus} agents produce localized approximations for the network average, \ie,  
\begin{equation}\label{eqn:limit} \lim_{k\to\infty}{x^{k}_i}=N^{-1}\sum^N_i x^{(0)}_i\end{equation} 
($N$ number of agents). Refer to Appendix~\ref{appendix:metropolis} for instructions on building such weights locally.

\textbf{Gradient tracking}: Enables agents to compute the solution $x$ of a distributed optimization problem. The distributed optimization problem arises on large scale machine learning/statistical learning problems~\citep{forero_2010}. Let every agent in $\N$ to hold a cost function $f_i(x):\RR^M\to \RR$ the objective of distributed optimization is to find $x$ that minimizes the average of all the functions
\begin{equation}\label{eqn:objective}
 \min_{x\in \RR^M} f(x)\triangleq \frac{1}{N}\sum^N_{i=1} f_i(x)
\end{equation}
using local communication and local computation. The algorithm for finding the minimizer $x$ starts from an arbitrary solution $x_i^{(0)}$ and the local variable $z^{(k)}_i$, which tracks the gradient of the function $f(x)$, is initialized by the local gradient at point $x_i^{(0)}$, \ie, $z_i^{(0)}=\nabla f_i(x_i^{(0)})$. The algorithm proceeds using the update~\citep{qu_2018}:
\begin{align}\label{eqn:GT} \begin{split}
x^{(k+1)}_i &= \sum^{N}_{j=1} \alpha_{i,j} x^{(k)}_j  - \eta z^{(k)}_i \text{,}\\
z^{(k+1)}_i &= \sum^{N}_ {j=1} \alpha_{i,j} z^{(k)}_j + \nabla f_i(x^{(k+1)}_i)  - \nabla f_i(x^{(k)}_i)\text{,}
\end{split}\end{align}
where $\left[\alpha_{i,j}\right]_{N\times N}$ are the consensus weights for a fixed strongly connected graph, and $\eta > 0$ is a fixed step size.

\textbf{Decentralized partially observable Markov decision process} (Dec-POMDP) is the framework describing the system dynamics where fully cooperative agents interact under partially observability settings--defined by the sextuple~\citep{gronauer_2022}:
$$\M=(\N, \S, \{\A_i\}_{i\in\N}, \{\O_i\}_{i\in\N}, \P, \bar{R}, \gamma)\text{,}$$
where $\N=\{1, \cdots, N\}$ denotes the set of interacting agents, $\S$ is the set of global but unobserved system states, and $\A_i$ is the set of individual action spaces. The observation space $\O$ denotes the collection of individual observations spaces $\O_i$. Typically an observation $o^t_i\in\O_i$ is a function of the state $s^t$. The state action transition probability function is denoted by $\P$, the team reward function shared by the agents $\bar{R}(s^t, a^t, s^{t+1})$, and the discount factor is $\gamma$. Agents observe $o^t_i\in\O_i$, choose an action from their individual action space $a_i\in \A_i$ and collect a common reward $\bar{R}(s^t, a^t, s^{t+1})$ . The system transitions to the next state following the state-action transition probability function. Neither the state $s^t$ or the joint action $a^t = \left[a^t_1, ..., a^t_N\right]$ is known to the agents.

The joint policy $\pi: \O \to  \Delta(\A)$ maps the joint observation $o^t$ to a distribution $\Delta$ over the joint action $\A = \A_1 \times \cdots \times \A_N$. The agents' objective is to find a joint policy $\pi$ that maximize the expected discounted return $J(\pi)$ over a finite horizon $T$ given by:

\begin{equation}\label{eqn:TDR}
    J(\pi)=\TDR{\sum^{T}_{t=0}\gamma^t  R(s^t, a^t, s^{t+1})}\text{.}
\end{equation}
Where $0< \gamma < 1$ is a discount factor that balances the preference between collecting immediate high rewards while avoiding future low rewarding states, and $\mu(s)$ is the initial state distribution. The joint reward $R(s^t, a^t, s^{t+1})$ is the sum of team rewards $\Sigma^N_i\bar{R}(s^t, a^t, s^{t+1})$. Thus the expected discounted return captures the expected sum of exponentially weighted joint rewards, by drawing an initial state  $s^0$ from $\mu$, observing $o^0$ and following the actions  prescribed by the joint policy $\pi$ thereafter until $T$.

\section{Distributed Value Decomposition}

Consider a setting where fully cooperative learners interact under a partially observable setting. Motivated by Fact~\ref{fac:VDN} which establishes that value decomposition networks minimize the mean squared {\em joint temporal difference} JTD.  We propose distributed value decomposition networks. DVDN use peer-to-peer message exchange to combine their local temporal difference (TD) for approximating JTD. Thus, agents emulate VDN weight updates using the JTD surrogate. Additionally, homogeneous agents share knowledge by pushing both weights and gradients to the communication channel, using gradient tracking. We formalize the setting whereby agents face uncertainty with respect to the communication channel with {\em hybrid partially observable Markov decision process} introduced by~\citet{santos_2024}.

\subsection{Preliminaries}

 \textbf{Value decomposition network \citep{sunehag_2018}}: is a {\em value-based} method for multi-agent reinforcement learning. In value-based methods, the expected discounted return in~\eqref{eqn:TDR} is captured by a joint $Q$-function that maps the joint observations $o$ and joint actions $a$ to a real value number ($Q$-value). Particularly, individual $Q$-functions (local observations and actions mappings to a $Q$-value) are estimated using parameterized non-linear function approximation, implementing the well known deep-$Q$ networ~\cite {mnih_2015} architecture in single agent RL. Then, in centralized training, agents combine their individual $Q$-functions into a joint $Q$-function: 
 
\begin{equation}\label{eqn:Q_VDN} Q^\text{VDN}(o, a; \omega) = \sum^N_{i=1} Q_i(o_i, a_i; \omega_i) \end{equation}
Where $\omega$ is the concatenation of the individual network parameters $\omega_i$, $o$ and $a$ represent the concatenation over the agents' observations and actions respectively. Finally, $Q^\text{VDN}$ is said to have {\em additive} factorization because it can be obtained directly by summing over agents' $Q$-function. The loss function used with additive factorization is:

\begin{equation}\label{eqn:loss_VDN}
\ell(\omega; \tau) = \frac{1}{N}\sum_\tau\left[y^\text{VDN} - Q^\text{VDN}(o, a;\omega)\right]^2\text{,}
\end{equation}
where the joint trajectory $\tau =[\tau_1,\tau_2,\cdots,\tau_N]$ is the concatenation of the trajectories drawn individually by interacting with the environment. The individual trajectory $\tau_i$ of episode with timesteps $t=0, ..., T$, is given by:
\begin{equation*}\tau_i = (o_i^0, a_i^0, \bar{R}^1, o_i^1, a_i^1, \bar{R}^2, \cdots, \bar{R}^T, o_i^T)\text{.}\end{equation*}
The parameter set $\omega=[\omega_1,\omega_2,\cdots,\omega_N]$ also factorizes across agents, such that each of the agent-wise policies $\pi^Q_i(o_i; \omega_i)$ is determined only by its $\omega_i$. Like deep $Q$-networks, VDN uses a target network which stabilizes training by providing the a joint target $y^\text{VDN}$:

\begin{equation} \label{eqn:y_VDN}
y^\text{VDN} = \sum^N_{i=1} \bar{R} + \gamma \max_{u_i}Q_i(o_i', u_i; \omega_i^-)\text{,} 
\end{equation}
where $o_i'$ is the subsequent observation to $o_i$ and $\omega_i^-$ is a periodic copy from $\omega_i$. Similar to the joint $Q$-function in~\eqref{eqn:Q_VDN} the joint target also factorizes into individual targets
\begin{equation}\label{eqn:y_i} y_i =  \bar{R} + \gamma \max_{u_i} Q_i(o'_i, u_i; \omega^-_i)\text{.}\end{equation}

\textbf{Hybrid partially observable Markov decision process} (H-POMDP) is the framework where agents cooperate under partial observability and are uncertain about the communication graph topology. During training, agents are unaware of whom their communicating peers are going to be at any given training episode. An extension from Dec-POMDPs, H-POMDP is defined by the sextuple~\citep{santos_2024}: 

$$\H=(\G, \S, \{\A_i\}_{i\in\N}, \{\O_i\}_{i\in\N}, \P, \bar{R}, \gamma)\text{,}$$
where $\G$  denotes an undirected {\em episode-varying communication graph} where  set of interacting agents remains fixed $\N$ but the communication edge set $\E$ is allowed to change between episodes. The other tuple elements adhere to the standard Dec-POMDP definition:  $\S$ is the set of global but unobserved system states, and $\A_i$ is the set of individual action spaces. The observation space $\O$ denotes the collection of individual observations spaces $\O_i$. Typically an observation $o^t_i\in\O_i$ is a function of the state $s^t$. The state action transition probability function is denoted by $\P$, the team reward function shared by the agents $\bar{R}(s^t, a^t, s^{t+1})$, and the discount factor is $\gamma$.

The unknown communication graph $\G$, is sampled from a set $\mathcal{C}$ according to an unknown probability distribution $\beta$. The agents performance is measured as $J_\beta(\pi) = \mathbb{E}_{\G\sim\beta}\left[ J(\pi; \G)\right]$, where $J(\pi; \G)$ denotes the expected discounted return of policy $\pi$ under an H-POMDP with communication graph $\G$. 

\subsection{Joint Temporal Difference}
    
Additive factorization is the distinguishing VDN characteristic; it is accomplished through a centralized value decomposition layer. The value decomposition layer sums the outputs from agents' deep $Q$-networks. And to emulate its behavior in the decentralized setting we must  determine: {\em  What information flows to agents' networks during centralized training?} To answer this research question, we  must first establish the role that value decomposition layer plays in shaping the weights of the $Q$-functions.

{\em Temporal difference} (TD) $\delta_i$ is the increment by which agent $i$ adjusts its $Q$-function in the direction of the maximal $Q$-value,  \ie,
\begin{equation}\label{eqn:TD}
\delta_i^t = \bar{R} + \gamma \max_{u_i}Q_i(o_i', u_i; \omega_i^-) -Q_i(o_i, a_i; \omega_i)\text{,} 
\end{equation} 
where the observation pair ($o_i$, $o_i'$) are surrogates for the current and next states respectively, in timestep $t$, the environment reaches hidden state $s^t$ and emits $o^t_i$ to agent $i$.

\begin{fact}\label{fac:VDN}
Value decomposition network minimize the joint temporal difference  $\delta=\Sigma_{j\in\N} \delta_j$, where
\begin{equation}\label{eqn:JTD}  \delta_j^t = \sum_\tau \bar{R} + \gamma \max_{u_j} Q_j( o_j', u_j; \omega_j^-) - Q_j( o_j, a_j;\omega_j)\text{.}\end{equation}
and $\delta_j\in\RR^T$ is agent $j$'s temporal difference, for the joint trajectory $\tau$ with length $T+1$.
\end{fact}

Fact~\ref{fac:VDN} establishes that the information flowing to weights through the back-propagation algorithm is the sum across agents of their temporal difference. Hence, in VDN, every agent replaces its own TD in~\eqref{eqn:TD} with JTD~\eqref{eqn:JTD}, for local updates. By construction, JTD decomposes between agents, and we use the fact that the joint trajectory in~\eqref{eqn:TD} is a concatenation of the individual trajectories $\tau_j$, to perform localized approximations to~\eqref{eqn:JTD}.

\subsection{Decentralized JTD}

{\em Can the effect of value decomposition layer be reproduced in the decentralized setting?} In the case where the communication graph is fully connected the system is back to the centralized setting and we can use VDN normally. Conversely, when communication is not possible the system is {\em fully decentralized}. {\em Independent $Q$-learners} perform weight updates in isolation by minimizing the mean square temporal difference: \begin{equation}\label{eqn:IL} \loss{\omega_i; \tau_i} := \frac{1}{T}\sum_{\tau_i}\delta_i^2\text{.}
\end{equation}

Alternatively, for situations where the communication graph is strongly connected, it is possible to generate a localized approximation for additive factorized joint $Q$-function from individual $Q$-functions. Since there is no agent capable of overseeing the system, we propose to perform the consensus updates in~\eqref{eqn:consensus}, to propagate the temporal difference: 
\begin{equation}\label{eqn:consensus_JTD} 
    \delta^{(k+1)}_i = \sum_{\JN} \alpha^{(k)}_{i, j} \delta^{(k)}_j\text{.}
\end{equation}
As a result from temporal difference consensus in~\eqref{eqn:consensus_JTD} agents receive an updated estimation for the team temporal difference. The {\em team temporal difference estimator} at communication step $k$ at agent $i$ is denoted by $\delta^{(k+1)}_i$. The limit in~\eqref{eqn:limit} guarantees that the updates asymptotically converge to $^1/_N\delta$. However, with a finite number of updates it is not possible to guarantee that the agents will reach consensus. Instead, we resort to truncation. In this work we perform a {\em single} consensus step per mini-batch update.

In a practical applications, it is useful to separate temporal difference (TD) from the additional information that comes from communication. The {\em network estimated JTD } at agent $i$ $\hat{\delta_i}$ captures the contributions that come from network and is defined as:
$$\hat{\delta}_{-i} = N\delta^{(k+1)}_i - \delta^{(k)}_i\text{.}$$
The network estimated JTD at agent $i$ can be used to perform weight updates in the direction that minimizes the mean square estimated JTD at agent $i$: 

\begin{equation}\label{eqn:DVDN} 
    \ell(\omega_i; \tau_i, \hat{\delta}_{-i} ) := \frac{1}{T}\sum_{\tau_i}\left(\delta_i + \hat{\delta}_{-i}\right)^2
\end{equation}
Since $\hat{\delta}_{-i}$ is the value obtained from the network of agents, it comes after the semicolon in the LHS of~\eqref{eqn:DVDN}, in the place reserved for data. Whereas the variable in the minimization is the $Q$-network parameters $\omega_i$. Similarly, $\delta_i$ is a variable which depends on the (trajectory, parameters) ($\tau_i, \omega_i$).  Figure~\ref{fig:diagram} (Appendix~\ref{appendix:method}) compares the error increments that shape the weight updates in VDN and DVDN in \eqref{eqn:consensus_JTD} and~\eqref{eqn:DVDN}. 
    
Differently from the distributed algorithm max-push~\citep{kok_2006},  where the communication is {\em serial} and it happens on an established order, induced by an spanning tree that is common knowledge, the updates in~\eqref{eqn:consensus_JTD} happen in {\em parallel}. Moreover, max-push requires the design of joint action utility functions, to capture the value of the joint action. Agents have a time budget to agree on what the best global action is by solving their local problems. There are no guarantees that max-push converges to a global joint action in graphs with cycles, or within their allocated time budget. The updates in DVDN have asymptotic convergence guarantees and do not require explicit joint action modeling, or even knowledge of neighbors' actions.

\subsection{DVDN with Gradient Tracking}

When the agents are homogeneous,\ie, they have the same observation and action space, it is possible to use gradient tracking to make them agree on a common solution for the parameters. This common solution combines weight updates from  many agents improving sample-efficiency. We assume that there is a global {\em loss function} that factorizes additively agent-wise and the solution is a common $\omega$. The global loss function can be expressed by replacing the loss function~\eqref{eqn:DVDN} in the agent wise cost functions in~\eqref{eqn:objective}:
\begin{equation}\label{eqn:GT_unconstrained}
    \min_\omega \ell(\omega; \tau) = \frac{1}{N}\sum^N_{i=1}\ell_i(\omega)
\end{equation}
Since the minimizer $\omega$ for the global in~\eqref{eqn:GT_unconstrained} is unknown to the agents, an alternative formulation introduces $N$ copies $\omega$ and followed by $N$ equality constraints: 
\begin{equation}
\begin{aligned}\label{eqn:GT_constrained}
    \min_{\omega_i\in\Omega, \forall i \in \N}\ell(\omega; \tau) &= \frac{1}{N}\sum^N_{i=1}\ell_i(\omega_i)\\
    \text{s.t. } x_i = x_j &\quad \quad\forall (i, j) \in \E 
\end{aligned}
\end{equation}
for the parameter space $\Omega$.  We rewrite the terms in~\eqref{eqn:GT_constrained} making explicit the dependency of the local loss functions in LHS with the local experiences in~\eqref{eqn:DVDN}:  
\begin{equation}
    \begin{aligned}\label{eqn:GT_objective}
    \min_{\omega_i\in\Omega, \forall i \in \N}\ell(\omega; \tau) &= \frac{1}{N}\sum^N_{i=1}\ell(\omega_i; \tau_i, \hat{\delta}_{-i})\\
    \text{s.t. } x_i = x_j &\quad \quad\forall (i, j) \in \E 
    \end{aligned}
\end{equation}

Similar to previous work~\citep{nedic_2020}, we propose a gradient-based strategy for solving~\eqref{eqn:GT_objective}. Differently from previous work in which the objective function is strongly convex, here the objective is non-convex. As result there is no closed form solution to the optimization problem.  Moreover, we cannot guarantee  that the equality constraints are satisfied since we truncate the number of consensus steps; Rather, we interleave gradient descent steps for minimization with consensus steps to incentivize an alignment in the solution. A practical  reinforcement learning algorithm is initialized as follows:

At the first mini-batch iteration $k=1$, agents perform a forward pass on  the local networks and compute $\delta^{(1)}_i$, then they perform the consensus iteration in~\eqref{eqn:consensus_JTD} to obtain the estimated network JTD at agent $i$,
$$
\hat{\delta}^{(1)}_{-i} = N\left(\sum_{j\in\N^{(1)}_i} \alpha^{(1)}_{i, j} \delta^{(1)}\right) - \delta^{(1)}_i\text{.}
$$
Then, agents compute the gradient of the mean square estimated JTD~\eqref{eqn:DVDN}
$$
g^{(1)}_i = \nabla_{\omega_i}\ell(\omega^{(0)}_i; \tau^{(1)}_i, \delta^{(1)}_{-i})
 $$
We introduce an auxiliary variable $z^{(1)}_i$ that tracks the team gradient 
$$\nabla \ell(\omega; \tau) = \frac{1}{N} \sum^N_{i=1} \nabla \ell(\omega_i; \tau_i)\text{,}$$ 
and is initialized at the local gradient loss, \ie,
$$z^{(1)}_i = g^{(1)}\text{.}$$
The update at step $k=1$, combines a consensus step~\eqref{eqn:consensus} and gradient descent with weight updates $z^{(1)}_i$: 
$$\omega^{(1)}_i =  \sum_{j\in\N^{(2)}_i} \alpha^{(1)}_{i, j} \omega^{(0)}_i -\eta z^{(1)}_i \text{.}$$
At the second mini-batch iteration $k=2$, agents compute the estimated network JTD at agent $i$,
$$
\hat{\delta}^{(2)}_{-i} = N(\sum_{j\in\N^{(2)}_i} \alpha^{(2)}_{i, j} \hat{\delta}^{(2)}_j) - \delta^{(2)}_i\text{.}
$$
Then, agents compute the local gradients at iteration $k=2$:
$$
g^{(2)}_i = \nabla_{\omega_i}\ell(\omega^{(1)}_i; \tau^{(2)}_i, \delta^{(2)}_{-i}).
 $$
Once difference between local gradients is available, agents update the team gradient variable:
 $$
 z^{(2)}_i = (\sum_{j\in\N^{(2)}_i} \alpha^{(2)}_{i, j} z^{(1)}_j) + g^{(2)}_i- g^{(1)}_i\text{.}
 $$
 Finally, agents aggregate the weights and perform a gradient step at iteration $k=2$ using the team gradient instead of the local gradients: 
 $$
 \omega^{(2)}_i = \sum_{j\in\N^{(2)}_i} \alpha^{(2)}_{i, j} \omega^{(1)}_j -\eta z^{(2)}_i
 $$
 ending the iteration $k=2$ updates. More generally, for an arbitrary $k$: 
 
\begin{subequations}\label{eqn:DVDN_GT} 
    \begin{align}
    \hat{\delta}^{(k)}_{-i} &= N\left(\sum_{j\in\N^{(k)}_i} \alpha^{(k)}_{i, j} \delta^{(k)}_j\right) - \delta^{(k)}_i\label{eqn:DVDN_GT:JTD}\\
     g^{(k)}_i &= \nabla \ell(\omega^{(k-1)}_i;\tau^{(k)}_i, \delta^{(k)}_{-i})\label{eqn:DVDN_GT:g}\\
     z^{(k)}_i &=  \sum_{\JN} \alpha^{(k)}_{i,j}z^{(k-1)}_i+ g^{(k)}_i - g^{(k-1)}_i\label{eqn:DVDN_GT:z}\\ 
     \omega^{(k)}_i &=  \sum_{\JN} \alpha^{(k)}_{i,j}\omega^{(k-1)}_i - \eta z^{(k)}_i\label{eqn:DVDN_GT:w}
    \end{align}
\end{subequations}

The intuition behind the algorithm~\eqref{eqn:DVDN_GT} is to update the auxiliary variables, {\em team gradient variables}, $z_i$ in the direction that minimizes LHS of~\eqref{eqn:GT_objective}. Over many iterations of $k$, the team gradient variables $z^{(k)}_i$ converge asymptotically to the average of individual gradients.  As established in the previous section, we only perform a single update per mini-batch--which is sufficient to emulate the effect of parameter sharing in the distributed setting, and provides better results than the fully decentralized independent learner. The practical implementation of this algorithm incorporates adaptive momentum (Adam)~\citep{kingma_2014} updates to improve its performance, making it compatible with standard VDN implementation.

\section{Experiments}

We evaluate the performance of both DVDN algorithms in ten scenarios with partial observability, and compare them to two baselines: The independent $Q$-learning in the fully decentralized paradigm, and value decomposition networks in the CTDE paradigm. IQL is the lower bound baseline and VDN is the upper bound baseline. Specifically, for each scenario, in the heterogeneous agents setting, we compare VDN to DVDN (Algorithm~\ref{alg:DVDN}, Appendix~\ref{appendix:DVDN}), and in the homogeneous agents setting, we compare VDN (PS) to DVDN (GT) (Algorithm~\ref{alg:DVDN_GT}, Appendix~\ref{appendix:DVDN_GT}).

\subsection{Scenarios}

We consider three environments: The first is {\em level-based foraging}~\citep[LBF,][]{papoudakis_2021}\footnote{\url{https://github.com/uoe-agents/lb-foraging}}, where agents collect fruits in a grid-world and are rewarded if their level is greater than or equal to the fruits they are  loading. Agents perceive the X-Y positions and levels of both food items and peers within a two block radius. Rewards are sparse and positive, they depend on the level of the fruit item being loaded and the level of each contributing agent. The second environment is the {\em multi-particle environment}~\citep[MPE,][]{lowe_2017} where agents navigate in a continuous grid, with their trajectories dependent on past movement actions. We modify the original environment for partial observability\footnote{\url{https://github.com/GAIPS/multiagent-particle-envs}}. The third environment~\citep[MARBLER,][]{torbati_2023} is a robotic navigation simulator that generates physical robot dynamics. In this environment, all agents are assumed to have full communication and their observations are appended together. However, agents remain aware only of their own actions. The scenarios configurations are:

\begin{itemize}
    \item {\textbf{LBF/Easy}}: Three agents interact in a 10x10 grid-world to collect 3 food items.
    \item {\textbf{LBF/Medium}}: Four agents interact in a 15x15 grid-world to collect 5 food items.
    \item {\textbf{LBF/Hard}}: Three agents interact in a 15x15 grid-world to collect 5 food items.
    \item {\textbf{MPE/Adversary}}: Two teammates must guard the target landmark against the approach of a pretrained adversary agent.  They perceive their own position, their relative distance to the goal, the positions of the landmarks, and the position and color of the closest agent. The color enables a teammate to distinguish whether or not the other agent is the adversary. The team is rewarded in proportion to the negative distance from the target to its closest teammate and penalized by the adversary's  distance to the target.
    \item {\textbf{MPE/Spread}}: Three agents must navigate as close as possible to one of three landmarks. They perceive their own position, velocity, the position and velocity of the closest agent and the position of the closest landmark. The team is rewarded the negative distance from the agent closest to a landmark  and receive a penalty for bumping into one another.
    \item {\textbf{MPE/Tag}}: Three large and slow moving predator agents must bump into a smaller and faster pretrained  prey agent. Predator agents perceive their own position and velocity, the position and velocity  of the closest predator, and the position and velocity of the prey. Predators collect rewards by touching the prey.
    \item{\textbf{MARBLER/Arctic Transport}} (Arctic):  Two drones attempt to guide the ice robot and water robot to the goal location as quickly as possible over ground tiles (white), ice tiles (light blue), and water tiles (dark blue). Drones move fast, while ground robots move fast on their specialized tiles. Teammates are penalized at each time step  in proportion to the number of the ground robots not in the destination, plus the distance of the ground robots from their goal.
    \item{\textbf{MARBLER/Material Transport}} (Material): A heterogeneous team of four agents, two fast agents  with low loading capacity, and two slow agents with high loading capacity, must collaborate to unload two zones. The teammates collect positive rewards for loading and unloading the zones, while collecting a small penalty per timestep.
    \item{\textbf{MARBLER/Predator Captures Prey}} (PCP): A heterogeneous team of two sensing agents and two capture agents must collaborate to capture six preys. A prey can be perceived only by one of the sensing agents, and can be captured only by one of the capture agents. Teammates collect positive rewards for sensing and capturing the prey, and a small penalty per timestep.
    \item{\textbf{MARBLER/Warehouse}}: A team of six robots, three green and three red, must navigate to the zone of their color on the right side of the environment to receive a load. Then they must navigate to their color zone on the left side to unload. Since optimal paths intersect robots must coordinate to avoid collisions. Teammates collect positive rewards for successfully loading and unloading the zones.
\end{itemize}

\begin{table*}
   \setlength{\tabcolsep}{.15em}
   \centering
    \caption{Maximum average episodic returns, over independent seeds,  their respective 95\%  bootstrapped confidence interval for all algorithms and tasks. Highlighted results are those with the higher maximum average episodic returns. The asterisk denotes results that match the performance of the best result for the task (see Section~\ref{sec:metrics}). The double asterisk denotes results that are second in the rank.}\label{tab:max}
    \begin{center}
        \begin{tabular}{clccccccc}
            \toprule
                                    & & \multicolumn{3}{c}{\textbf{Heterogeneous}} & &\multicolumn{2}{c}{\textbf{Homogeneous}}\\\cline{3-5}\cline{7-9}
               \textbf{Env.} & \textbf{Task.} &  \textbf{IQL} & \textbf{DVDN } & \textbf{VDN} &  & \textbf{IQL} & \textbf{DVDN (GT)} & \textbf{VDN (PS)}\\
             \midrule
             \parbox[t]{2mm}{\multirow{5}{*}{\rotatebox[origin=c]{90}{LBF}}} 
                     & Easy & \CI{0.81}{-0.02, 0.02} &  \CI{0.80}{-0.02,0.02} & \CB{0.85}{-0.02,0.02} & & \CI{0.81}{-0.02, 0.02} & \CI{0.89^{**}}{-0.02, 0.02} & \CB{0.94}{-0.01, 0.01}\\ 
                    & Medium &  \CI{0.61}{-0.02, 0.02} & \CI{0.62^{**}}{-0.02,0.02}& \CB{0.64}{-0.02,0.02}& & \CI{0.61}{-0.02, 0.02} & \CI{0.72^{**}}{-0.01, 0.02} & \CB{0.79}{-0.02, 0.02}\\
                    & Hard & \CI{0.43^{*}}{-0.01, 0.02} &  \CB{0.44}{-0.02, 0.02}  & \CB{0.44}{-0.02, 0.02} & & \CI{0.43}{-0.01, 0.02} & \CI{0.52^{**}}{-0.01, 0.02} & \CB{0.56}{-0.02, 0.02}\\\hline
             \parbox[t]{2mm}{\multirow{5}{*}{\rotatebox[origin=c]{90}{MPE}}} 
                    & Adversary & \CI{9.26^{**}}{-0.57, 0.52} &  \CB{10.05}{-0.48, 0.43} & \CI{8.82}{-0.43, 0.40} & & \CB{9.26}{-0.56,0.52} & \CI{8.66}{-0.27, 0.30} & \CI{8.73}{-0.42, 0.44}\\
                    &  Spread &  \CB{-135.81}{-1.74, 1.68} & \CI{-137.44^{*}}{-1.98, 1.98} & \CI{-144.05}{-1.62, 1.61} & & \CI{-135.81^{**}}{-1.74, 1.68} & \CB{-131.62}{-1.52, 1.55} & \CI{-141.45}{-1.46, 1.36}\\
                    & Tag & \CI{23.00^{**}}{-1.74, 1.80} &  \CB{31.41}{-2.68, 2.98}  & \CI{18.33}{-1.40, 1.36} & & \CI{23.00^{*}}{-1.70, 1.78} & \CI{15.99}{-4.43, 4.13} & \CB{23.180}{-1.61, 1.68}\\\hline
             \parbox[t]{2mm}{\multirow{7}{*}{\rotatebox[origin=c]{90}{MARBLER}}} 
                    & Arctic & \CI{-43.51}{-1.64, 1.65} &  \CI{-37.56^{**}}{-1.01, 0.84} & \CB{-30.93}{-0.70, 0.76} & & \CI{-43.51^{**}}{-1.64,1.67} & \CI{-49.50}{-3.50, 3.03} & \CB{-28.55}{-0.84, 1.11}\\
                    &  Material &  \CI{12.81}{-0.49, 0.51} & \CI{18.07^{**}}{-1.14, 1.30} & \CB{21.82}{-0.36, 0.36} & & \CI{12.81}{-0.49, 0.50} & \CI{17.32^{**}}{-0.83, 0.87} & \CB{21.94}{-0.56, 0.53}\\
                    & PCP & \CI{130.72^{**}}{-0.81, 0.76} &  \CB{133.02}{-0.67, 0.78}  & \CI{125.10}{-2.57, 3.09} & & \CB{130.72}{-0.86, 0.75} & \CI{124.36}{-1.85, 2.21} & \CI{125.27}{-1.46, 1.50}\\
                    & Warehouse & \CI{21.99}{-0.42, 0.38} &  \CB{28.74}{-0.45, 0.45}  & \CI{23.65^{**}}{-0.90, 0.93} & & \CI{21.99}{-0.43, 0.38} & \CB{29.54}{-2.36, 2.21} & \CI{28.79^{*}}{-0.57, 0.60}\\
            \bottomrule
        \end{tabular}
    \end{center}
    \vspace*{-2ex}
\end{table*} 

\subsection{Baselines}

We apply the algorithmic implementations of  IQL and VDN in~\citep{papoudakis_2021}\footnote{\url{https://github.com/uoe-agents/epymarl}} for the fully cooperative setting with joint rewards:

\begin{itemize}
    \item {\textbf{VDN}}~\citep{sunehag_2018}: Is the CTDE paradigm baseline. For the homogeneous agents setting,  we perform parameter sharing.
    \item {\textbf{IQL}}~\citep{mnih_2015}: Implement individual deep-$Q$ network that minimize a local loss function~\eqref{eqn:IL}. Since IQL belongs to the fully decentralized approach, we only perform experiments without parameter sharing for this algorithm.
\end{itemize}
    
\subsection{Metrics}\label{sec:metrics}

\textbf{Performance plots}: Our evaluation methodology follows that of ~\citet{papoudakis_2021}. We conduct ten independent, randomly seeded, runs  for each algorithm-scenario combination for five million training steps. Evaluations are performed at every 125,000 timesteps, for a total of forty one evaluation checkpoints over each training run. Each evaluation checkpoint consists of hundred independent episodes per training seed. The {\em average episodic return} is the average of the ten evaluation checkpoints (\ie, average over seeds). The {\em maximum average episodic return} captures the average episodic returns with maximum value (\ie, maximum over forty one average episodic returns).  We report the maximum average episodic return and the 95\% bootstrapped confidence interval (CI). 

\textbf{Ablation Plots}: To measure the contributions of different factors to performance gains, we conduct an ablation study. For each factor, we identify the evaluation checkpoint with the maximum average episodic return (over five independent seeds) and extend the selection to include a two-evaluation checkpoint neighborhood. This extended selection increases the group's sample size from a total of five data points to twenty-five, thereby decreasing standard deviation. We randomly resample this sample (size 20,000) to build a 95\% bootstrap confidence interval (CI). We report the resampled mean and bootstrap CI bars.

\textbf{Ranking criteria}:  To compare algorithms' performances, we refrain from using statistical language since their performance is based on the maximum evaluation checkpoints for a small number of independent runs, and may not necessarily follow any parametric distribution. Instead, to discriminate the algorithms' performances, we use a bootstrap CI test\footnote{\url{https://github.com/facebookincubator/bootstrapped}}. We say that algorithm A {\em matches} the performance of algorithm B when we cannot reject the null hypothesis that their distributions have the same mean.  When the bootstrap CI test rejects the null hypothesis that both algorithms have the same mean:  We say that algorithm A {\em outperforms} algorithm B if its average is greater. Otherwise,  we say that algorithm A {\em under performs} algorithm B.

\subsection{Setup}

For the baselines, we use the default hyper parameters from~\citep{papoudakis_2021}, but extend the number of training steps from two million to five million. For both DVDN algorithms, we perform hyper parameter search using the protocol in~\citep{papoudakis_2021}. To generate the switching topology network, we make a random sample from all possible strongly connected graphs \footnote{In the distributed optimization literature, strongly connected graphs are used in fixed topology network studies. While the term {\em switching topology} usually applies to settings where agents might be disconnected for some consensus iterations. We deviate from this norm by requiring that every agent be connected at every iteration.} for the task-specific number of agents. This sampling scheme ensures  that there is always a path between two agents. The hyper parameters used in the experiments are in the Appendix~\ref{appendix:hyperparameters}. Open-source code is available at the DVDN Github repository.\footnote{\url{https://github.com/GAIPS/DVDN}}.

\section{Results}\label{sec:results}

\begin{figure*}[!tb]
    \centering
    \begin{subfigure}{0.27\linewidth}
        \includegraphics[width=\linewidth]{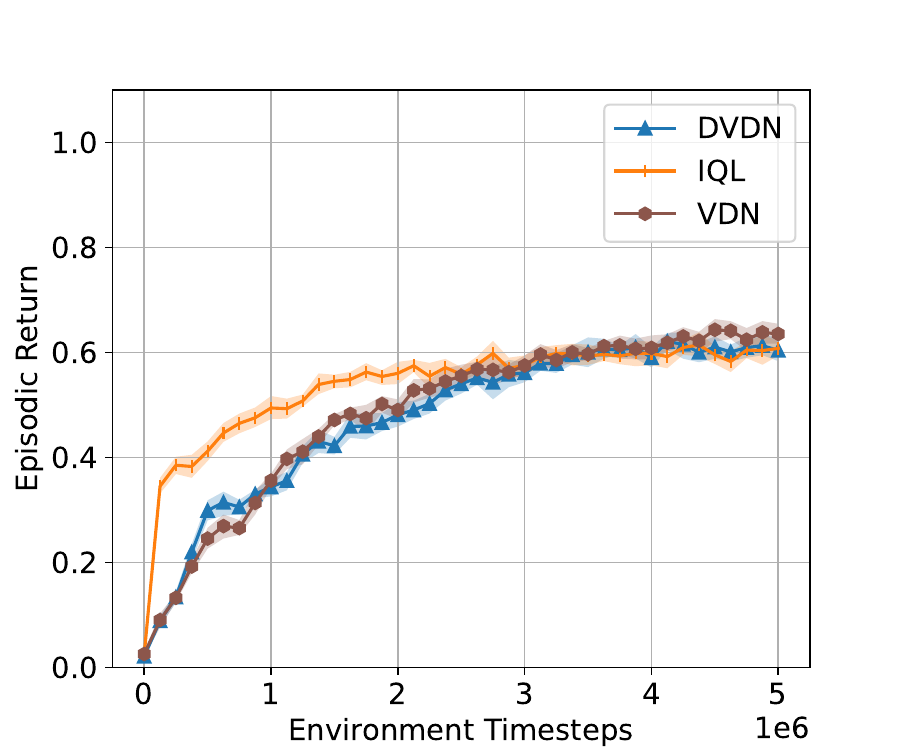}
        \caption{LBF (Medium)}
    \end{subfigure}\hfill
    \begin{subfigure}{0.27\linewidth}
        \includegraphics[width=\linewidth]{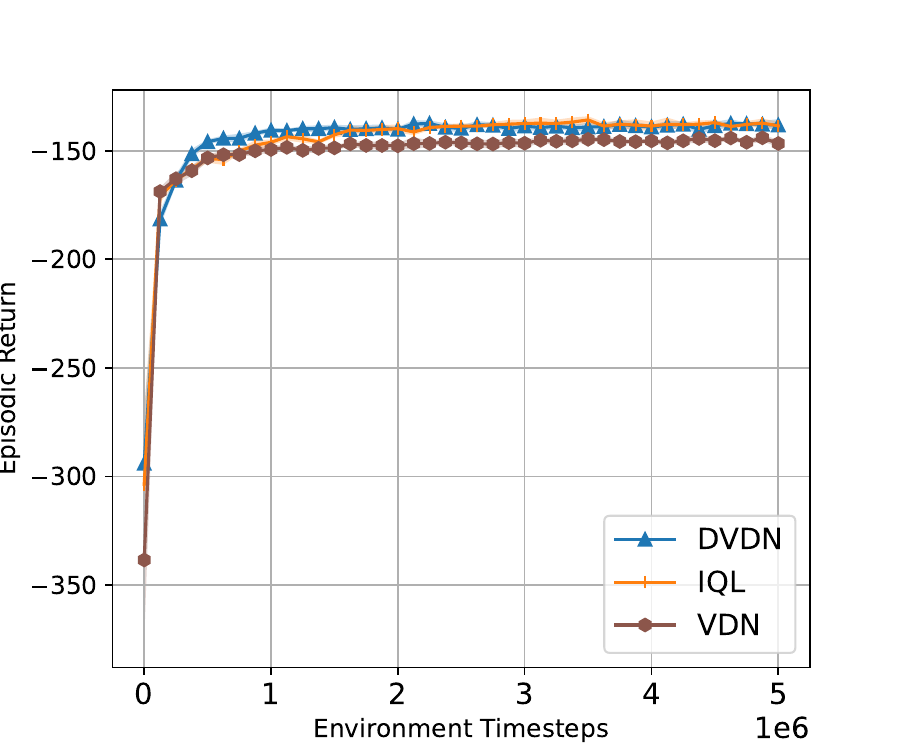}
        \caption{MPE (Spread)}
    \end{subfigure}\hfill
    \begin{subfigure}{0.27\linewidth}
        \includegraphics[width=\linewidth]{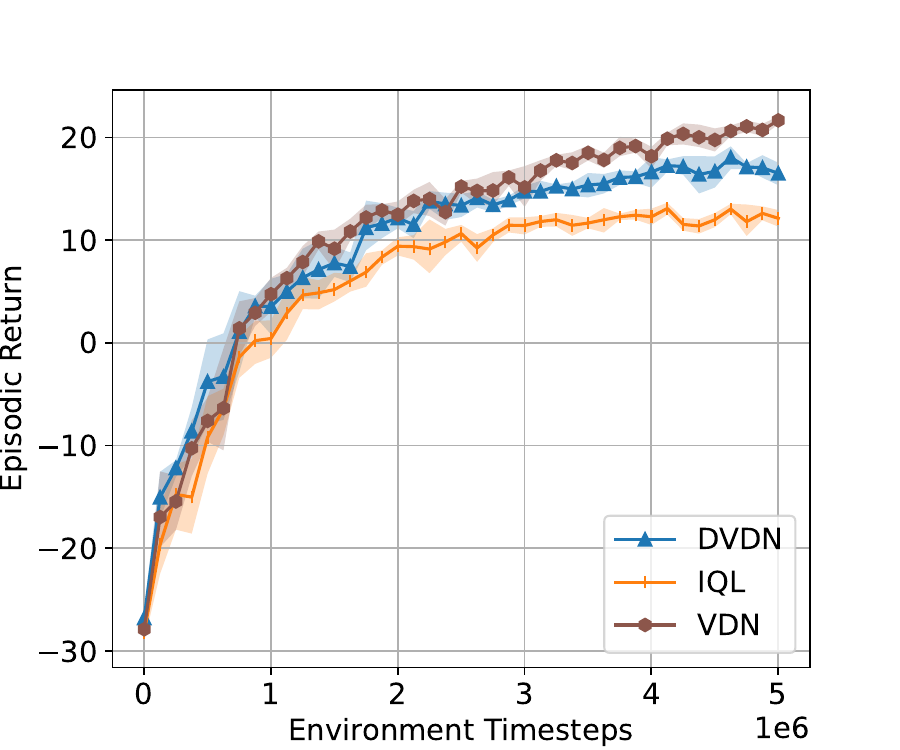}
        \caption{MARBLER (Material)}
    \end{subfigure}
    \begin{subfigure}{0.27\linewidth}
        \includegraphics[width=\linewidth]{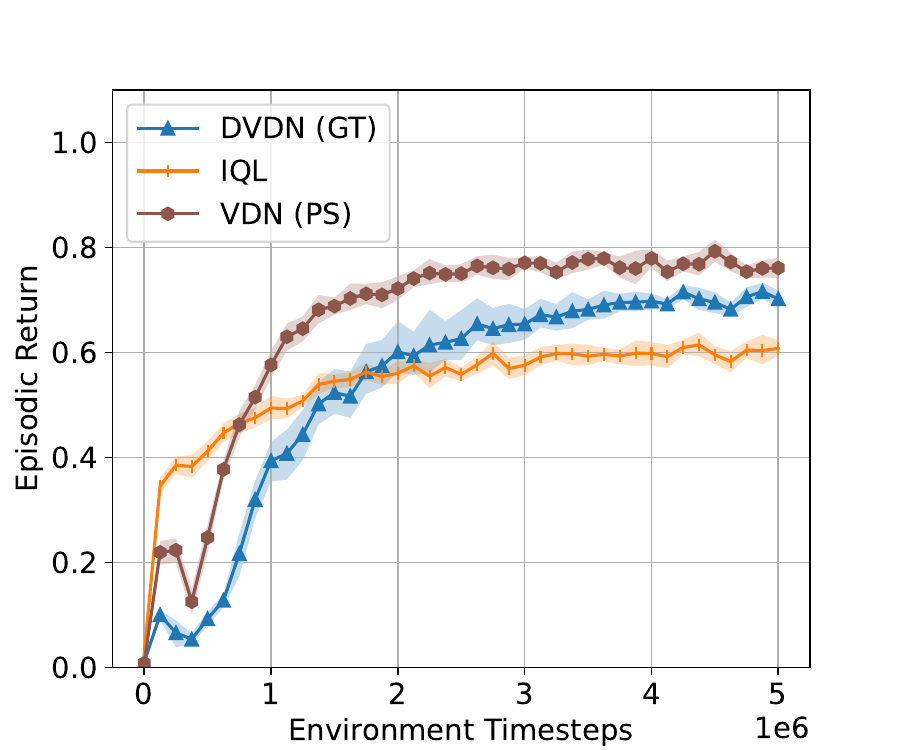}
        \caption{LBF (Medium)}
    \end{subfigure}\hfill
    \begin{subfigure}{0.27\linewidth}
        \includegraphics[width=\linewidth]{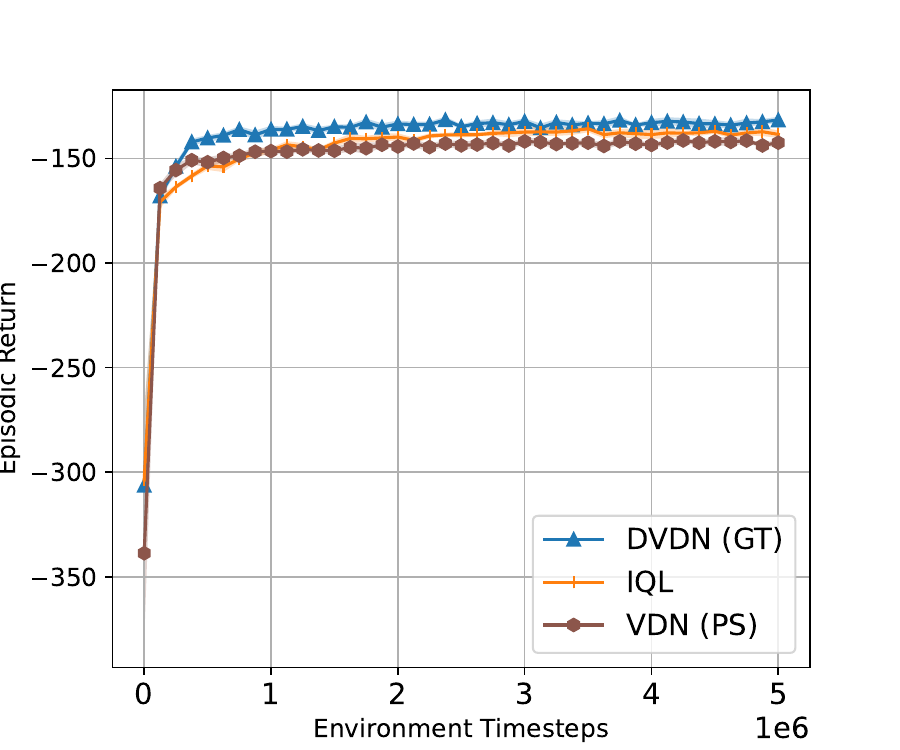}
        \caption{MPE (Spread)}
    \end{subfigure}\hfill
    \begin{subfigure}{0.27\linewidth}
        \includegraphics[width=\linewidth]{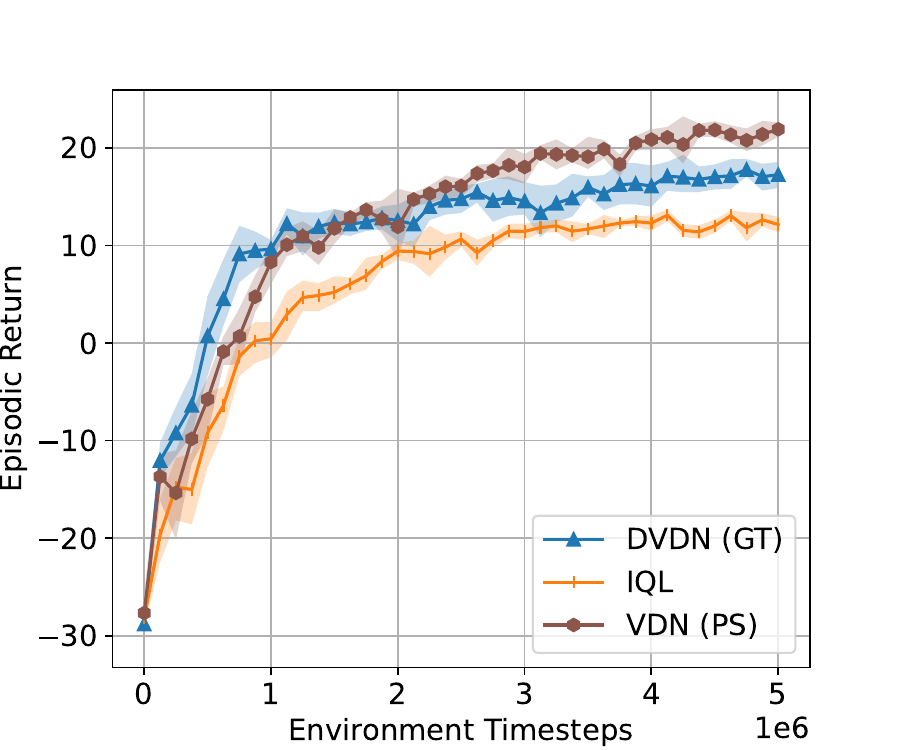}
        \caption{MARBLER (Material)}
    \end{subfigure}
    \caption{In the columns, the results for each environment is expressed by a representative task. In the top row, heterogeneous agents setting and in the bottom row homogeneous agents setting. The IQL curve is orange, VDN curve is chestnut and the DVDN curve is blue. The markers represent the evaluation checkpoint and the shaded area represent the 95\% bootstrap CIs. Notably, the performance curves for VDN and DVDN are similar, showcasing the effectiveness of using JTD as training signal.}\label{fig:test}
    \vspace{-4ex}
\end{figure*}

Figure~\ref{fig:test} illustrates the performance curves for one representative scenario per environment. The performance curve profiles are similar between VDN and DVDN. Table~\ref{tab:max} summarizes the results for IQL, DVDN and VDN algorithms, separating them into settings for heterogeneous agents and homogeneous agents. According to our ranking criteria, highlighted results indicate the best for the algorithm-scenario combination, asterisks mark performances that comparable to the best performing, and double asterisks indicate second-best performance. We argue that DVDN is an alternative to VDN, not its replacement. Consequently, it is sufficient to either outperform the lowest-ranked algorithm or match the performance of the highest-ranking algorithm. For the heterogeneous setting, DVDN matches or outperforms VDN in six of ten scenarios. For the homogeneous setting, DVDN (GT) outperforms IQL in five of ten scenarios. 

In LBF scenarios where the reward is sparse, agents in the homogeneous setting that apply parameter sharing or gradient tracking outperform their counterparts in heterogeneous setting. This result is consistent with previous authors~\citep{papoudakis_2021} claims, that larger grid-worlds (and/or, fewer agents) scenarios, benefit from parameter sharing. In PS, the same set of weights is updated using trajectories of all agents. In this environment, the benefits of using GT correlate with those of using PS.

In general, scenarios in the MPE environment show that IQL generates effective policies that are difficult to surpass by using value decomposition. Moreover, there is little performance differences between algorithms, even when comparing between the heterogeneous and homogeneous settings. The exception is the Tag scenario, where DVDN outperforms both baselines in the heterogeneous setting by a wide margin. Since DVDN produces localized approximations for the JTD, it produces policies that are more diverse, thus during exploration phase it experiences a wider variety of states. In fact, DVDN's best-performing policy has two predators fleeing from the prey, causing the prey to stay still, while the third predator bumps repeatedly into it. Because the prey moves according to a pretrained policy, DVDN learns a way to exploit the prey's policy by generating unseen states during the prey's training. For further discussion on this matter, please refer to Appendix~\ref{appendix:qualitative_analysis}.

Scenarios in the MARBLER environment showcase heterogeneous agents, sparse rewards, and agents are assumed to learn from the concatenation of observations.  In the heterogeneous setting,  DVDN's performance stands out, surpassing IQL in four scenarios and VDN in two scenarios. Whereas in the homogeneous setting, DVDN (GT)'s lags behind DVDN's. These findings contrast with those of the LBF environment, where GT enhances DVDN's performance. This performance degradation is anticipated, given that three out of four scenarios involve heterogeneous agents. However, further improvement is possible by performing GT in subsets of homogeneous teammates within the team. 

\begin{figure}[!tb]
    \centering
    \begin{subfigure}{0.49\linewidth}
        \centering
        \includegraphics[width=\linewidth]{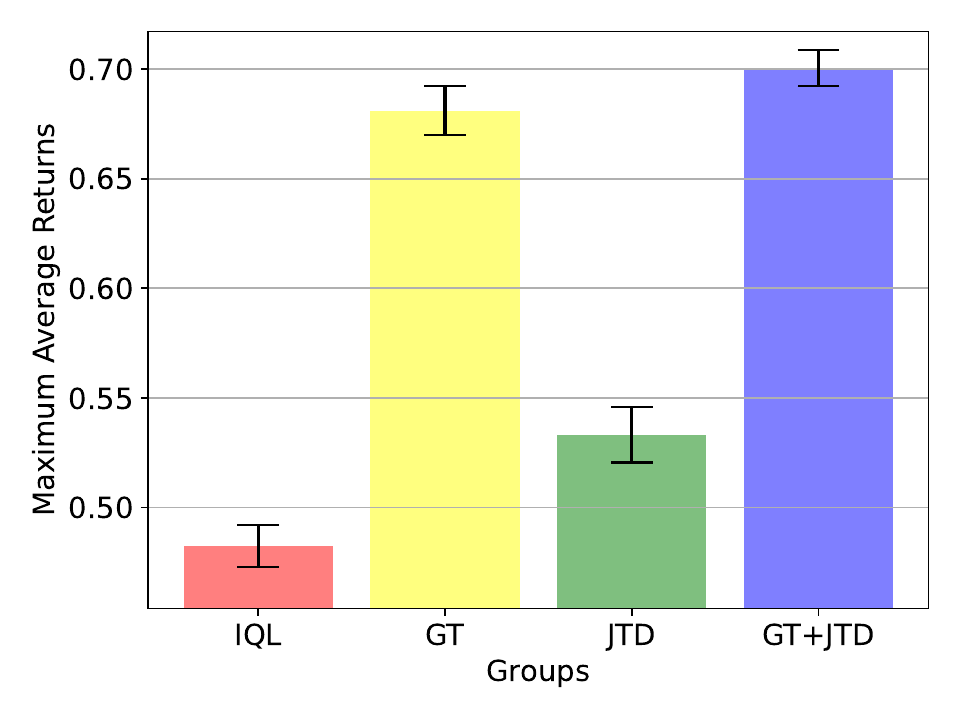}
        \caption{LBF (Medium)}
    \end{subfigure}\hfill
    \begin{subfigure}{0.49\linewidth}
        \centering
        \includegraphics[width=\linewidth]{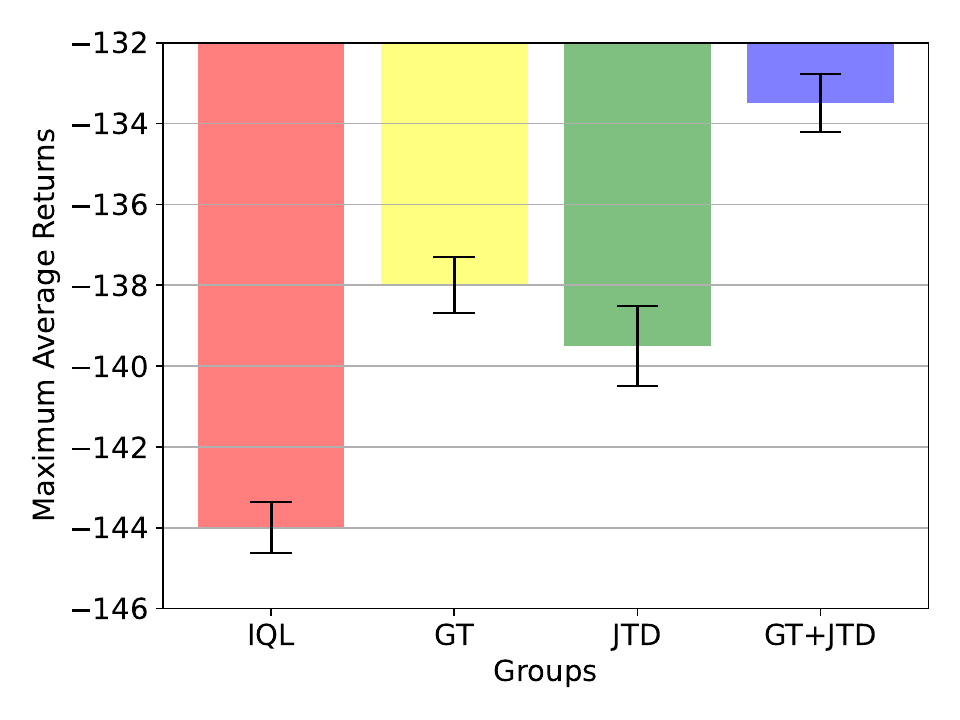}
        \caption{MPE (Spread)}
    \end{subfigure}\hfill
    \begin{subfigure}{0.49\linewidth}
        \centering
        \includegraphics[width=\linewidth]{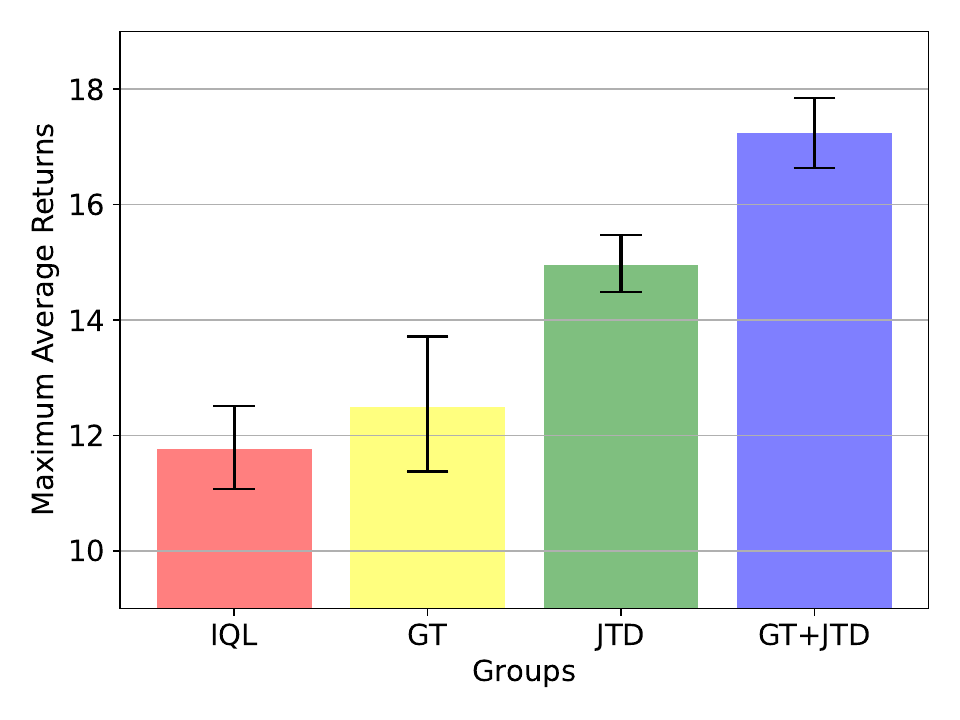}
        \caption{MARBLER (Material)}
    \end{subfigure}\hfill
    \caption{Ablation plots for the homogeneous setting, for the LBF and MARBLER environments respectively. The IQL group has no consensus (control group). The GT group performs gradient tracking. The JTD group performs joint temporal difference consensus. The GT+JTD group combines GT and JTD consensus. For the three tasks both factors individually improve results and are better combined.}\label{fig:ablation}
\end{figure}

The ablation plots (Fig.~\ref{fig:ablation}) illustrate the key factors contributing to DVDN's algorithmic performance for each environment and in a specific task. Starting with the base DVDN (GT) configuration used for testing, (GT+JTD),  we toggle on and off the JTD and GT modules. The IQL group (control group) has no communication, updating its weights by minimizing the loss as specified in~\eqref{eqn:IL}.  The GT group performs gradient tracking exclusively. The JTD group performs JTD consensus updating the loss in~\eqref{eqn:DVDN}.

For the three tasks in Fig.~\eqref{fig:ablation} both JTD and GT demonstrate improvements over IQL group, with the combined factors (GT+JTD) performing even better. This result ratify reported results in Table~\ref{tab:max}; Gradient tracking plays a crucial role in improving the performance for LBF Medium task. While joint temporal difference is the main factor enhancing the MARBLER Material task performance. In the MARBLER environment, GT slightly deteriorates the performance, while JTD improves it. The combined GT+JTD performance is about equivalent to the JTD group. For the MPE Spread task, both factors can individually enhance performance and combining them results in superior performance.

\section{Related Work}
 
\textbf{Distributed optimization}: The set of problems in decentralized decision-making where each agent holds a different  but related piece of information, and makes compatible decisions traces back to~\citet{tsitsiklis_1985}. These problems are characterized by local communication and local gradient minimization of a shared (convex) cost function~\citep{nedic_2020}.The network averaging consensus algorithm for the fixed strongly connected networks~\citep{xiao_2003}, and for switching topology networks~\citep{xiao_2007}, are essential components for one-hop communication between agents; They enable their agreement on a solution vector to the common decision problem. Large scale convex (supervised) machine learning problems can be within this decentralized  optimization framework, \eg,~\citep{forero_2010}~\citep{chang_2020}, where agents agree on a solution vector while maintaining the input data private. Gradient tracking~\citet{qu_2018}, performs consensus both on the solution vector and its local gradients, minimizing an objective that is the average of all local objectives. However, those agents are not reinforcement learning based agents. In reinforcement learning the state changes according to a transition function.

\textbf{Distributed optimization and deep learning}: A recent development is the investigation of distributed optimization in large scale supervised deep learning, \eg, ~\citep{jiang_2017} develop a image detection algorithm using consensus where each agent has a \textit{i.i.d} sample of a very large dataset. The focus on non-convex problems includes rigorous convergence analysis and the application of momentum based optimizers for better sample efficiency. In the centralized setting, momentum based optimizers, such as the Adam optimizer~\citet{kingma_2014}, boost the performance of deep learning models by incorporating the weight update history. This improvement carries over to the distributed setting, where consensus on model weights is integrated with momentum-based weight updates. For instance, consensus adaptive optimizer presented in~\citet{nazari_2022} surpasses the performance of  centralized optimizers. More recently,~\citet{carnevale_2023} propose gradient tracking Adam optimizer, GTAdam. Their system combines gradient tracking and adaptive momentum, and outperforms  other distributed optimizers. In supervised deep learning, the datasets are static whereas in reinforcement learning the datasets are dynamic. They change in accordance to shifts in agents' policies.

\textbf{Distributed MARL}: A research thread in MARL applies consensus-based mechanisms to the distributed {\em policy evaluation} problem. The problem of estimating the total discounted joint reward under a fixed joint policy under the fully observable setting. Agents perform localized approximations for the state value function, which can either be tabular or approximated using linear function approximation. Under those assumptions, there are methods that guarantee the convergence of the policy evaluation. Examples include:  DTDT~\citep{wang_2020}, Decentralized TD(0) with gradient tracking~\citep{lin_2022},  and~\citep{sun_2020}. Moving to the partially observable setting, consensus-based mechanisms also improve the performance of belief based agents.~\citet{kayaalp_2023} propose policy evaluation, with linear function approximation, under the partially observable setting whereby agents use the consensus mechanism to share both policies' parameters and beliefs with closest neighbors.~\citet{petric_2023} proposes learning agents that perform consensus on a tabular belief, and learn polices using interior point methods. In this work, we propose to use non-convex function approximation under partially observability.

\textbf{Coordination graphs} (CGs): is  a solution to coordination problems in games where agents interact with a few neighbors to decide on a local joint action; Agents are nodes and joint actions are edges representing a local coordination dependency. Agents learn the {\em payoff} function for every joint action without exploring the large combinatorial action space. The global payoff functions is the sum of local payoff functions. For solving the CG, ~\citet{guestrin_2002} propose a variable elimination algorithm. However, this approach  can scale exponentially in the number of agents for densely connected graphs~\citep{kok_2006}.~\citet{kok_2006} propose the distributed {\em max-plus} algorithm, that out performs  variable elimination for densely connected graphs. However, the underlying CG must be fixed during training, it induces a spanning tree that models the communication channel, and it requires a variable number of message exchanges so that all agents agree on a the best (local) payoff. Moreover, convergence of this message passing scheme is only guaranteed for acyclic CGs. Both variable elimination and max-plus algorithm have been developed to work on the tabular payoff function case. Deep coordination graphs~\cite[DCGs,][]{bohmer_2020} generalize max-plus algorithm to train end-to-end CGs, using parameter sharing between payoff functions and privileged information (global state). DCGs are expressive  enough to represent a rich set of $Q$-functions that factorize, such as VDN and QMIX~\cite[QMIX,][]{rashid_2018}. Particularly, all the methods assume centralized training and a fixed coordination graph topology which is common knowledge throughout training.

\textbf{Networked agents with multi-agent reinforcement learning}:~\citet{zhang_2018}  develops the first reinforcement learning-based agents with local communication, gradient minimization with asymptotic convergence guarantees using linear function approximation. This method, networked agents, promotes {\em decentralized training} and {\em decentralized execution} paradigm where agents learn locally and are suitable for real world infrastructure and robot teams domains~\citep{gronauer_2022}. However, there are three limitations to their work: the assumption of full state and action space observability, the slower sample efficiency due to stochastic online learning~\citep{mnih_2016} and the risk of convergence to a sub optimal Nash-equilibrium \citep{zhang_2021}.~\citet{chen_2022} extend networked agents for a mild form of partial observability--jointly observable state, where the state is fully observable taking into account all the of agents' perceptions~\citep{oliehoek_2016}. Their system adapts a well known CTDE algorithm, MADDPG~\citep{lowe_2017}, to the DTDE approach, allowing agents to use consensus iterations on weights to emulate parameter sharing. Our approach extends previous work, by  developing networked agents under partial observability and applying gradient tracking  as means of performing localized weight updates of a shared objective.

\section{Conclusion and Future Work}
    
We contribute with {\em distributed value decomposition network} DVDN whereby agents learn by performing consensus on temporal difference. DVDN replicates the gradient updates of  VDN in the decentralized setting using communication. We investigate the effect of the learning signal in two settings, heterogeneous and homogeneous agents. Homogeneous agents also align their policies' parameters using gradient tracking. We implement our approach in ten scenarios with partial observability with favorable results. Our approach is suitable to settings where there is no centralized node.  For future work we intend on incorporating belief sharing to DVDN, and the extension of other families of factorized $Q$-function such as QMIX~\citep{rashid_2018} to the distributed setting.



\bibliographystyle{ACM-Reference-Format} 
\bibliography{dvdn}


\newpage
\appendix
\onecolumn
\pagenumbering{roman}

\setcounter{page}{1}
\section{Extended Background}\label{appendix:metropolis}

The metropolis weights associated to an arbitrary graph $\G(\N, \E)$ is given by~\citep{xiao_2007}: 
 \begin{equation}\label{eqn:metropolis}
      \alpha_{n, m}=
       \begin{cases}
         \alpha_{n, n} = 1 - \sum_{m\neq n} \alpha_{n, m}  & \quad \text{if $n=m$} \\
        \frac{1}{1 + \text{max}\left(d(n), d(m)\right)} & \quad \text{if $(n,m)\in\E$}\\
        0 & \quad \text{otherwise}
       \end{cases}
 \end{equation}
 Where $d(n)$ is the degree of nodes $n$, and $\E$ the edge set of the graph.  In a distributed setting, each agent $n$ communicates its degree only to its neighbors. Then, each agent $n$ can determine its weight $\alpha_{n, m}$ for each of its neighbors $m\in \N$.  After these preliminary computations, agents are free to perform consensus.

\section{Proof}\label{appendix:proof}

Additive value decomposition is computed by summing the $Q$-values from each agents' $Q$-functions. Therefore, the error increment at the output layer of the $Q$-functions, following value decomposition, equals the sum of individual temporal differences. To demonstrate Fact~\ref{fac:VDN}, we proceed starting with VDN's loss function, then we perform differentiation at the value decomposition layer. Finally, we show that the error increment at the output layer of an arbitrary agent $i$, following differentiation, equals the joint temporal difference. 

\begin{align}\label{eqn:VDN_expanded}
\ell(\omega; \tau) &=  \frac{1}{N}\sum_\tau\left[y^\text{VDN} - Q^\text{VDN}(o, a;\omega)\right]^2\quad\text{from~\eqref{eqn:loss_VDN}}\nonumber\\
                                      &= \frac{1}{N}\sum_\tau\left[ \sum^N_{i=1} r + \gamma \max_{u_i}Q_i(o_i', u_i; \omega_i^-) - \sum^N_{i=1} Q_i(o_i, a_i; \omega_i)\right]^2\quad\text{from~\eqref{eqn:y_VDN} and~\eqref{eqn:Q_VDN}}\nonumber\\
                                      &= \frac{1}{N}\sum_\tau\left[ \sum^N_{i=1} y_i - \sum^N_{i=1} Q_i(o_i, a_i; \omega_i)\right]^2\quad\text{from~\eqref{eqn:y_i}}\nonumber\\
                                      &= \frac{1}{N}\sum_\tau\left[ \sum^N_{i=1} y_i - \sum^N_{i=1} q_i\right]^2\quad\text{define $q_i := Q_i(o_i, a_i; \omega_i)$}\nonumber\\
                                    &= \frac{1}{N}\sum_\tau\left[\sum^N_{i=1}\big(y_i - q_i\big)\right]^2\nonumber\\
                                    &=\frac{1}{N}\sum_{\tau}\Big\{\sum^N_{i=1}\big(y_i - q_i\big)^2  + 2\sum^{N-1}_{i=1}\sum^{N}_{j>i}\big(y_i - q_i\big)\big(y_j - q_j\big)\Big\}\text{.}
\end{align}
By taking the derivative of the loss in \eqref{eqn:VDN_expanded} with respect to $\omega$:

\begin{equation}\begin{split}\label{eqn:VDN_gradient}
\frac{\partial}{\partial \omega}\ell(\omega; \tau) &=  -\frac{2}{N} \sum_{\tau}\Big\{\sum^N_{i=1}\big(y_i - q_i\big)\nabla_\omega q_i  +  \sum^{N-1}_{i=1}\sum^N_{j> i}\big[\big(y_i - q_i\big)\nabla_\omega q_j+\big(y_j - q_j\big)\nabla_\omega q_i\big]\Big\}\\
&=  -\frac{2}{N} \sum_{\tau}\Big\{\sum^N_{i=1}\big(y_i - q_i\big)\nabla_\omega q_i  +  \sum^{N-1}_{i=1}\sum^N_{j> i}\big(y_i - q_i\big)\nabla_\omega q_j+\sum^{N-1}_{j=1}\sum^N_{i> j}\big(y_i - q_i\big)\nabla_\omega q_j\Big\}\\
&=  -\frac{2}{N} \sum_{\tau}\Big\{\sum^N_{i=1}\big(y_i - q_i\big)\nabla_\omega q_i  + \sum^N_{i=1}\sum^N_{j\neq i}\big(y_j - q_j\big)\nabla_\omega q_i\Big\}\\
                        &= -\frac{2}{N}\sum_{\tau}\sum^N_{i=1}\left[\sum^N_{j=1}(y_j - q_j)\right]\nabla_\omega q_i \\
                        &= -\frac{2}{N}\sum_{\tau}\sum^N_{i=1}\big(\sum^N_{j=1}\delta_j\big)\nabla_\omega q_i\\
                        &= -\frac{2}{N}\sum_{\tau}\sum^N_{i=1}\delta\nabla_\omega q_i
\end{split}\end{equation}
The joint temporal difference $\delta=\Sigma_j\delta_j$ \eqref{eqn:VDN_gradient} represents the sum of local temporal difference errors across agents. Using the fact that agent-wise $Q$-functions factorize, making the local weight updates dependent on individual weights $\omega_i, i\in\N$. The local weight update at agent $i$, is a term from~\eqref{eqn:VDN_gradient}: 
\begin{align}\label{eqn:solution}
    \frac{\partial}{\partial \omega_i}\ell(\omega_i; \tau) &= -\frac{2}{N}\sum_{\tau}\delta\nabla_\omega q_i=-\frac{2}{N}\sum_{\tau}\delta\nabla_{\omega_i} q_i\quad\quad\quad\quad\quad\quad\square
\end{align}
Since the weight updates are computed sequentially from the output layer to inner layers (\citep{wichert_2021}), the term in~\eqref{eqn:solution} consists of the product of a constant factor $\frac{2}{N}$, a network term $\delta$ and the gradient of inner layers $\nabla_{\omega_i} q_i$. Since the gradients of the inner layers depend on local information--we complete the demonstration. 

\section{Extended Method}\label{appendix:method}

\begin{figure}[H]
    \centering
    \begin{subfigure}{0.40\linewidth}
        \includegraphics[width=\linewidth]{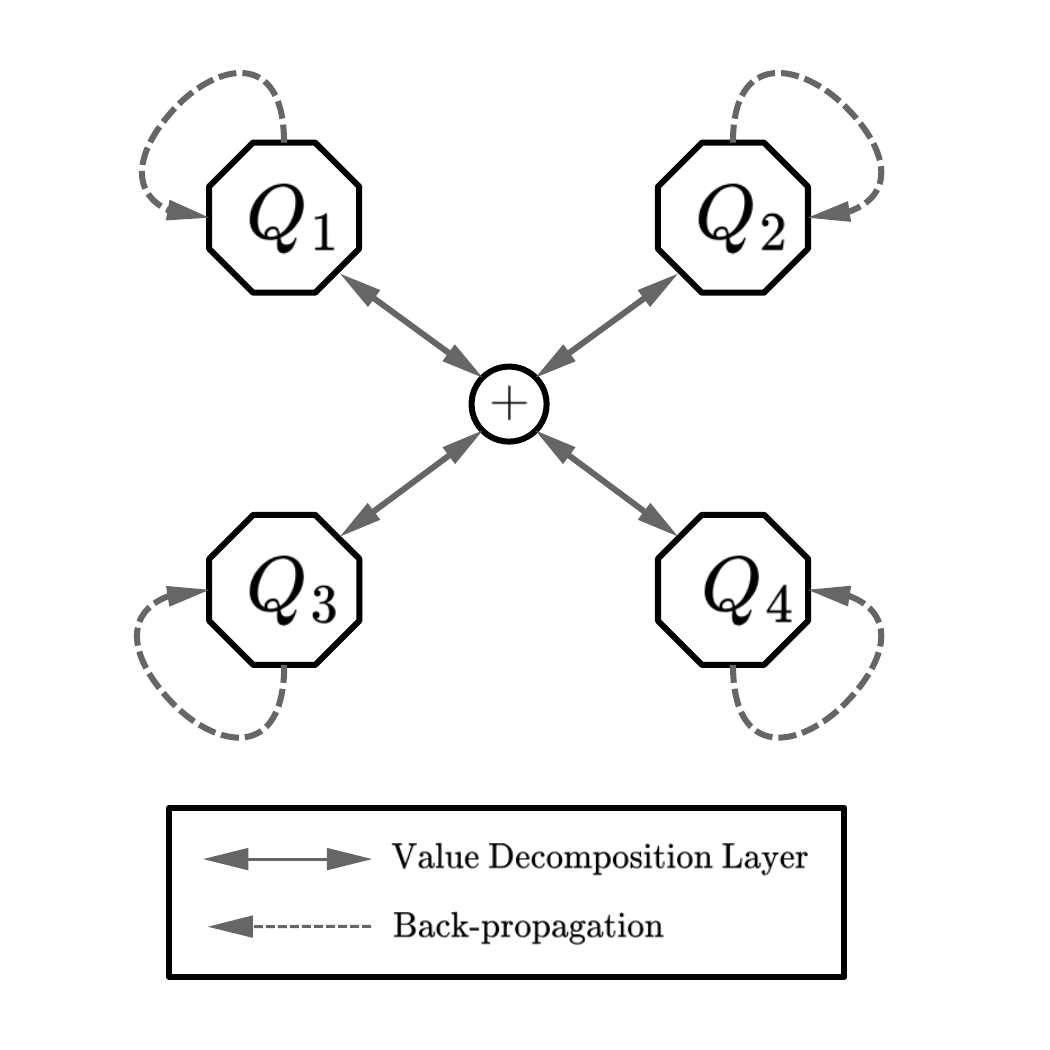}
        \caption{Value decomposition}
    \end{subfigure}
    \begin{subfigure}{0.40\linewidth}
        \includegraphics[width=\linewidth]{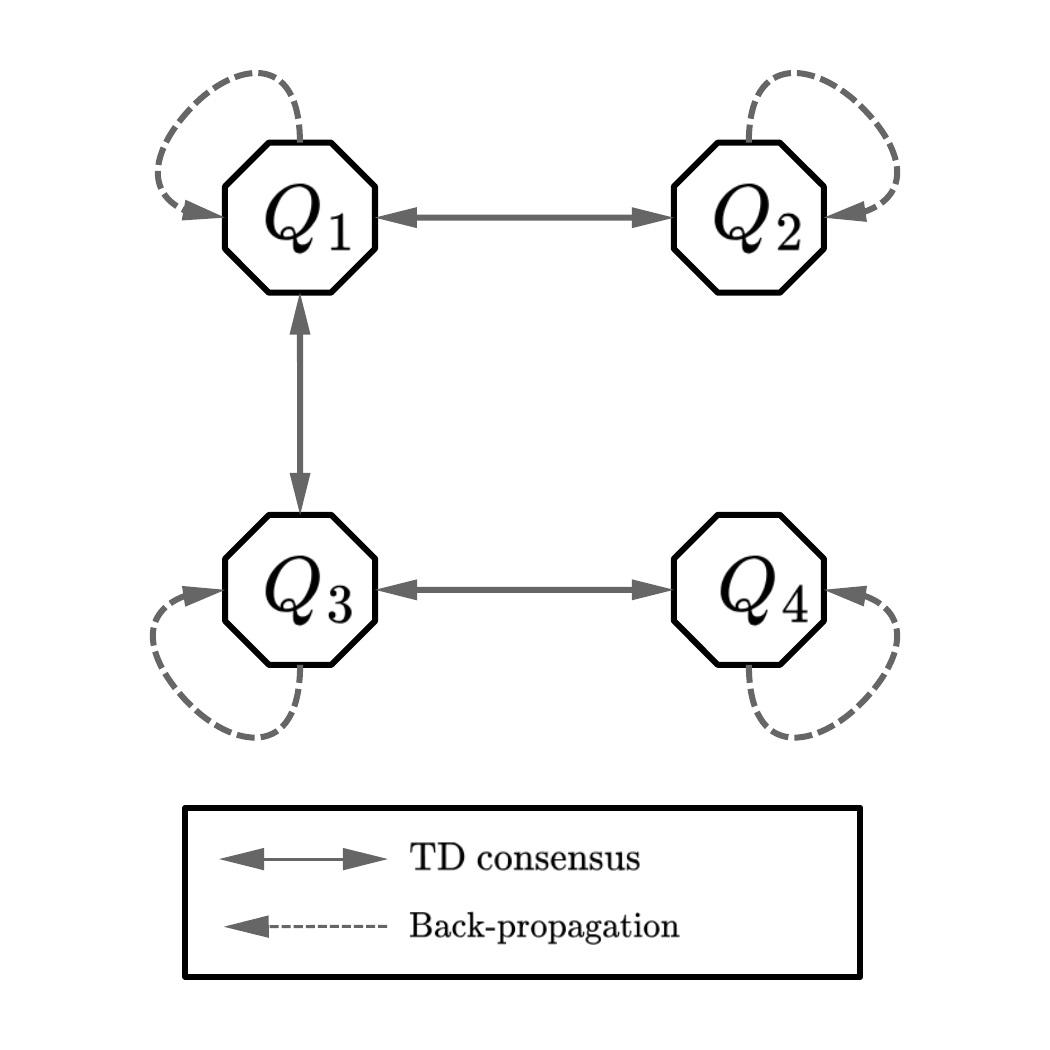}
        \caption{TD consensus}
    \end{subfigure}
    \caption{To the left, the value decomposition  diagram  where the value decomposition layer performs addition. To the right, the temporal difference consensus agents user {\em peer-to-peer} communication over an arbitrary strongly connected graph. Bidirectional arrows indicate the last step of forward pass (a) or TD consensus (b). The dashed line represent back-propagation algorithm.}\label{fig:diagram}
    \vspace{-3ex}    
\end{figure}

Figure~\ref{fig:diagram} contrasts the value decomposition sum operation (a) with temporal difference consensus (b) in~\eqref{eqn:consensus_JTD}. In Figure~\ref{fig:diagram} (a) Bidirectional arrows indicate the last step of forward pass: Agents contribute with their temporal difference and receive the joint temporal difference and back-propagation resumes. In Figure~\ref{fig:diagram}  (b) Bidirectional arrows indicate the temporal difference consensus step over an arbitrary communication graph.  Following consensus iteration agents perform back-propagation using the error increment $\delta_i+\hat{\delta}_{-i}$ (dashed arrow).

\section{Extended Experiments}

This section  presents pseudo-codes for distributed value decomposition networks with and without gradient tracking, and includes the hyper parameters for each version of the algorithm  as well as  baselines.  For simplicity, the code listings do not include an important code level optimization: {\em mini-batch updates}. Mini-batch updates accelerate learning by providing extra trajectories for the optimizer. Those extra trajectories are stored in a replay buffer, and are sampled randomly at each parameter update. On a distributed system the sampled trajectories have to be aligned to the same episode for the consensus rounds to succeed.  However, implementing this synchronization scheme is straightforward; every agent starts with the same long list of random replay buffer indices--it is their common knowledge. They draw the trajectories by sequentially consulting the batch size indices from this list, obtaining the same trajectories from the replay buffer.

 \subsection{Distributed VDN}\label{appendix:DVDN}
 
The distributed value decomposition networks in Listing~\ref{alg:DVDN} is a serial implementation of a synchronous distributed algorithm.  Agents communicate with their closest neighbors and have no knowledge of other teammates. It consists of three parts:  initialization, local communication and local gradient minimization. 
 
The $Q$-function parameters $\omega^{(0)}_i$ are initialized randomly, the Adam optimizer parameters~\citep{kingma_2014} are a constant learning rate $\eta$ , moving average parameter $\beta_1$, moving sum of squares parameter $\beta_2$, and a numerical stability parameter $\epsilon$ . Typical choices for hyper parameters are $\beta_1=0.9$, $\beta_2=0.999$  and $\epsilon=10^{-8}$.

The outer loop variable $k$ (lines 1-8) controls both the number of training episodes and consensus rounds. In parallel, agents collect individual trajectories $\tau_i$ (line 2). Agents compute temporal difference by performing a forward pass on their $Q$-networks (line 3). A new communication graph $\G^{(k)}$ is randomly generated at each episode $k$ as a strongly connected graph, ensuring the algorithm is not dependent on the network topology. Agents then exchange degree information with neighbors to derive the consensus weights (line 4). Next, they perform a single iteration of consensus to obtain an estimation of the joint temporal difference $\hat{\delta}^{(k)}_{-i}$ (line 5). The gradient is calculated based on the DVDN equation~\eqref{eqn:DVDN} (line 6). Finally, the weight $\omega^{(k)}_i$ is adjusted using the gradient and adaptive momentum weights~\footnote{This is a standard step for weight updates. Refer to~\texttt{\url{https://pytorch.org/docs/stable/generated/torch.optim.Adam.html}}} (line 7).
  
 \begin{algorithm}[H]
     \caption{Distributed Value Decomposition Network}\label{alg:DVDN}
     \begin{algorithmic}[1]
         \REQUIRE $\omega^{(0)}_i\text{ arbitrary}, \eta, \beta_1=0.9$, $\beta_2=0.999,\epsilon=10^{-8}$,  $\text{AdamOpt}_i = \text{Adam($\eta, \beta_1, \beta_2, \epsilon$)}$ 
          \FOR{$k=1,\cdots$}
                \STATE Observe $\tau_i\text{,}\, \forall i\in \N$
                \STATE $\delta^{(k)}_i =  \bar{R} + \gamma \max_u Q_i( o_i', u;\omega_i^-) - Q_i(o_i, a_i;\omega^{(k-1)}_i)$
                \STATE  Hear communication channel  and compute $\alpha^{(k)}_{i,j} \forall j\in \N$ using~\eqref{eqn:metropolis}
                \STATE $\hat{\delta}^{(k)}_{-i} = N \big(\sum_{\JN} \alpha^{(k)}_{i, j} \delta^{(k)}_j\big) - \delta^{(k)}_i$
                \STATE $g^{(k)}_i=\nabla_{\omega_i}\ell(\omega^{(k-1)}_i; \tau_i, \hat{\delta}^{(k)}_{-i})\quad\text{(Gradient of \eqref{eqn:DVDN})}$
                \STATE $\omega^{(k)}_i = \text{AdamOpti}_i.Step(\omega^{(k-1)}_i, g^{(k)}_i)$
            \ENDFOR
     \end{algorithmic}
 \end{algorithm}

\subsection{DVDN with Gradient Tracking}\label{appendix:DVDN_GT}

Homogeneous agents are interchangeable, having identical observation and action sets. In  centralized training, multiple agents share a single network parameterized by $\omega$ to enable efficient learning from a distributed set of experiences~\citep{gupta_2017}. In decentralized training, due to truncated consensus iterations,  agents may not have the same copy of parameters.   We combine the ideas of Algorithm~\ref{alg:DVDN} with gradient tracking~\eqref{eqn:GT}. Our approach utilizes previous work,~\citet{carnevale_2023}   that integrated gradient tracking with Adam optimizer. Listing~\ref{alg:DVDN_GT}  reports a serial implementation of the synchronous distributed value decomposition networks with gradient tracking (DVDN  (GT)):

 \begin{algorithm}[H]
     \caption{Distributed Value Decomposition Networks  With Gradient Tracking}\label{alg:DVDN_GT}
     \begin{algorithmic}[1]
         \REQUIRE $\omega^{(0)}_i\text{ arbitrary}, \eta, \beta_1=0.9$, $\beta_2=0.999,\epsilon=10^{-8}$, $\text{AdamOpt}_i= \text{Adam($\eta, \beta_1, \beta_2, \epsilon$)}$ 
          \FOR{$k=1,\cdots$}
                \STATE Observe $\tau_i\quad \forall i\in \N$
                \STATE $\delta^{(k)}_i =  \bar{R} + \gamma \max_u Q_i( o_i', u;\omega_i^-) - Q_i(o_i, a_i;\omega^{(k-1)}_i)$
                \STATE  Hear communication channel  and compute $\alpha^{(k)}_{i,j} \forall j\in \N$ using~\eqref{eqn:metropolis}
                \STATE $\hat{\delta}^{(k)}_{-i} = N \big(\sum_{\JN} \alpha^{(k)}_{i, j} \delta^{(k)}_j\big) - \delta^{(k)}_i\quad~\eqref{eqn:DVDN_GT:JTD}$
                \STATE $g^{(k)}_i=\nabla_{\omega_i}\ell(\omega^{(k-1)}_i; \tau_i, \hat{\delta}^{(k)}_{-i}) \quad\text{~\eqref{eqn:DVDN_GT:g}}$
                \IF{$k=1$}
                    \STATE $z^{(1)} = g^{(1)}_i\quad\text{ and }\quad\tilde{\omega}^{(1)}_i = \omega^{(0)}_i$
                \ELSE
                    \STATE $z^{(k)}_i =  \sum_{\JN} \alpha^{(k)}_{i,j}z^{(k-1)}_j +  g^{(k)}_i - g^{(k-1)}_i\quad\text{~\eqref{eqn:DVDN_GT:z}}$
                    \STATE $\tilde{\omega}^{(k)}_i = \sum_{\JN}\alpha^{(k)}_{i,j}\omega^{(k-1)}_j\quad\text{~\eqref{eqn:DVDN_GT:w}}$
                \ENDIF
                \STATE $\omega^{(k)}_i = \text{AdamOpt}_i.Step(\tilde{\omega}^{(k)}_i, z^{(k)}_i)$
            \ENDFOR
     \end{algorithmic}
 \end{algorithm}

 Algorithm~\ref{alg:DVDN_GT} requires each agent to hold two internal states: The previous local gradient update  $g^{(k)}_i=\nabla_{\omega_i}\ell(\omega^{(k-1)}_i;\tau_i, \hat{\delta}^{(k)}_{-i})$, and the previous team gradient update $z^{(k)}_i$ that is a local estimation to:
 \begin{equation}\label{eqn:network_gradient} 
 \nabla_\omega\ell(\omega;\tau, \delta) \approx \frac{1}{N}\sum_{i\in\N}\nabla_{\omega_i}\ell(\omega^{(k-1)}_i;\tau_i, \hat{\delta}^{(k)}_{-i})\text{.}
 \end{equation}
 
 Moreover, DVDN (GT), requires three consensus steps for every training episode: 

 \begin{itemize}
     \item  The network estimated JTD $\hat{\delta}^{(k)}_{-i}$ used in producing a local estimation ($\delta^{(k)}_i + \hat{\delta}^{(k)}_{-i}$)  to the joint temporal difference $\delta^{(k)}$ (line 5).
     \item  The team gradient $z^{(k)}_i$ approximates the left-hand side of ~\eqref{eqn:network_gradient} (lines 8, 10).
     \item   The parameter consensus aligns parameter estimations:  ${\omega}^{(k)}_i\approx\omega^{(k)}$ (line 11).
 \end{itemize}
 
The execution flow of the algorithm proceeds as follows:

The outer loop variable $k$ (lines 1-13) controls the number of training episodes and consensus rounds. In parallel, agents collect individual trajectories $\tau_i$ (line 2), and compute temporal differences by performing a forward pass on their $Q$-networks (line 3). A new communication graph  $\G^{(k)}$ is randomly generated at each episode $k$ as a strongly connected graph, ensuring the algorithm is not dependent on the network topology. Agents exchange degree information with neighbors to derive the consensus weights (line 4). Next, they perform a single iteration of consensus to obtain an estimation of the joint temporal difference $\hat{\delta}^{(k)}_{-i}$ (line 5). The gradient is calculated based on the DVDN equation (line 6). 

The first episode initializes the team gradient $z^{(1)}_i$ and consensus parameter $\tilde{\omega}^{(1)}$ (line 8).  Subsequent episodes update the team gradient by performing one consensus iteration on $z^{(k-1)}_i$ and adding the difference between the current and previous local gradient (line 10). Agents also perform one consensus iteration on the parameters $w^{(k-1)}_i$ (line 11).  Finally, the weight $\omega^{(k)}_i$ is adjusted using the gradient and adaptive momentum weights (line 13).

 \subsection{Hyperparameters}\label{appendix:hyperparameters}

 We conduct a grid search over a range of hyper parameter values for three environments. Table~\ref{tab:hypersearch} details hyperparameters values and descriptions. The hyperparameters of the DVDN algorithms are optimized for one scenario within each environment and remain consistent across other scenarios within the same environment. The selected scenarios for hyperparameter tuning are LBF medium instance, Tag for the multi-particle environment, and Simple Navigation for MARBLER. Each combination of hyperparameter is evaluated for three seeds, with the optimal hyperparameters being those that yield the maximum average return across the seeds. For LBF and MPE environments, baseline algorithms were not optimized and their hyperparameter configurations are as stated in~\citep{papoudakis_2021}. In contrast, the IQL and VDN baselines in the MARBLER environment have their hyperparameters optimized through our grid search. The VDN (PS) configuration remains unchanged from~\citep{torbati_2023}. Finally, the hyperparameter search process used two million timesteps for environments LBF and MPE, and four hundred thousand timesteps for MARBLER. 
 
The selected hyperparameters are reported in Table~\ref{tab:LBF_MPE_ns_hyper} and  Table~\ref{tab:LBF_MPE_gt_hyper}  for the heterogeneous agents and homogeneous agents settings respectively for environments LBF and MPE. Table~\ref{tab:MARBLER_hyper} reports the selected hyperparameters for both settings for the MARBLER environment.

 \begin{table}[!tb]
    \centering
     \caption{Hyperparameters used for hyperparameter search~\citep{papoudakis_2021}.}\label{tab:hypersearch}
    \begin{tabular}{|>{\centering\arraybackslash}m{30mm}|m{80mm}|>{\centering\arraybackslash}m{30mm}|}
    \toprule
    \textbf{Hyperparameter}    & \multicolumn{1}{>{\centering\arraybackslash}m{80mm}|}{\textbf{Description}}   & \textbf{Values} \\
    \hline
      hidden dimension & The number of neurons in the hidden layer of the neural networks. & 64/128 \\
    \hline
      learning rate & Regulates the step size of the gradient updates. & 0.0001/ 0.0003/ 0.0005 \\
    \hline
      reward standardization & Performs reward normalization. & True  \\
    \hline
      network type & Feed forward and fully connected (FC) or Gated Recurrent Unit (GRU). & FC/GRU \\
    \hline
        target update &  In 0.01 (soft) mode the {\em target network} is updated with parameters from {\em behavior network} every training step, following a exponentially weighted moving average, with innovation rate 0.01. In 200 (hard) mode the target network is updated with a full copy from the behavior  network at every 200 training steps. &
                                  200 (hard)/ 0.01 (soft)\\
    \hline
         evaluation epsilon & Epsilon is a hyperparameter controlling the sampling of sub-optimal actions from $Q$-value based policies. The {\em epsilon greedy} criteria provides a way for the agent to experiment with non-greedy actions  actions to find more profitable states. Evaluation epsilon regulates the rate of non-greedy actions taken during evaluation checkpoints. & 0.0/0.05  \\
    \hline
         epsilon anneal &   The number of episode steps to reach the minimum epsilon for $Q$-value based policies. & 125,000/500,000 (LBF), 125,000/500,000 (MPE), 50,000/200,000 (MARBLER). \\
    \bottomrule
    \end{tabular}
\end{table}

\begin{table}[!tb]
    \caption{Selected hyperparameters for the heterogeneous agents setting for environments Level-based foraging and Multi-particle environment. We adopt the hyperparameters from  ~\protect\citep{papoudakis_2021} for the baselines (without parameter sharing) .}\label{tab:LBF_MPE_ns_hyper}
    \begin{center}
        \begin{tabular}{lrrrrrrr}
            \toprule
                                    & \multicolumn{3}{c}{\textbf{LBF}} & &\multicolumn{3}{c}{\textbf{MPE}}\\\cline{2-4}\cline{6-8}
               \textbf{Hyperparameter} &\textbf{IQL} & \textbf{DVDN} & \textbf{VDN} &  & \textbf{IQL} & \textbf{DVDN} & \textbf{VDN} \\
             \midrule
             hidden dimension & 64 & 128 & 64 & & 128 & 128 & 128\\
             learning rate & 0.0003 & 0.0001 & 0.0001 & & 0.0005 & 0.0003 & 0.0005\\
             reward standardization & True & True &  True & & True & True & True \\
             evaluation epsilon & 0.05 & 0.05 &  0.05 & & 0.0 & 0.0 & 0.0 \\
             network type & GRU & GRU & GRU & & FC & GRU & FC \\
             epsilon anneal time & 250,000 & 500,000 & 500,000 & & 500,000 & 500,000  & 125,000 \\
             target update &  200 (hard)& 0.01 (soft) & 200 (hard) & & 0.01 (soft) & 200 (hard) & 200 (hard) \\
            \bottomrule
        \end{tabular}
    \end{center}
\end{table}

\begin{table}[!tb]
    \caption{Selected hyperparameters for the homogeneous agents setting. We adopt the hyperparameters from  ~\protect\citep{papoudakis_2021} for the baselines (with parameter sharing).}\label{tab:LBF_MPE_gt_hyper}
    \begin{center}
        \begin{tabular}{lrrrrrrr}
            \toprule
                                    & \multicolumn{3}{c}{\textbf{LBF}} & &\multicolumn{3}{c}{\textbf{MPE}}\\\cline{2-4}\cline{6-8}
               \textbf{Hyperparameter} &\textbf{IQL} & \textbf{DVDN (GT)}  & \textbf{VDN (PS)} &  & \textbf{IQL} & \textbf{DVDN (GT)}  & \textbf{VDN (PS)} \\
             \midrule
             hidden dimension & 64 & 128 & 128 & & 128 & 128 & 128\\
             learning rate & 0.0003 & 0.0003 & 0.0003 & & 0.0005 & 0.0005 & 0.0005\\
             reward standardization & True & True &  True & & True & True & True \\
             evaluation epsilon & 0.05 & 0.00 &  0.00 & & 0.0 & 0.0 & 0.0 \\
             network type & GRU & GRU & GRU & & FC & GRU & FC \\
             epsilon anneal time & 250,000 & 500,000 & 500,000 & & 500,000 & 125,000  & 125,000 \\
             target update &  200 (hard)& 0.01 (soft) & 0.01 (soft) & & 0.01 (soft) & 200 (hard) & 200 (hard) \\
            \bottomrule
        \end{tabular}
    \end{center}
\end{table}

\begin{table}[!tb]
    \caption{Selected hyperparameters for MARBLER environment. We adopt the hyperparameters from  ~\protect\citep{torbati_2023} for VDN.}\label{tab:MARBLER_hyper}
    \begin{center}
        \begin{tabular}{lrrrrrr}
            \toprule
                                    & \multicolumn{3}{c}{\textbf{Heterogeneous}} & &\multicolumn{2}{c}{\textbf{Homogenous}}\\\cline{2-4}\cline{6-7}
               \textbf{Hyperparameter} &\textbf{IQL} & \textbf{DVDN} & \textbf{VDN} &  & \textbf{DVDN (GT)} & \textbf{VDN (PS)} \\
             \midrule
             hidden dimension & 128 & 128 & 64 & &  128 & 128\\
             learning rate & 0.0005 & 0.0005 & 0.0005 & & 0.0005 & 0.0003\\
             reward standardization & True & True &  False & &  True & False \\
             evaluation epsilon & 0.00 & 0.00 &  0.00 & &  0.05 & 0.0 \\
             network type & GRU & GRU & GRU & &  GRU & GRU \\
             epsilon anneal time & 50,000 & 50,000 & 50,000 & & 50,000  & 50,000 \\
             target update &  200 (hard)& 0.01 (soft) & 200 (hard) & & 0.01 (soft) & 200 (hard) \\
            \bottomrule
        \end{tabular}
    \end{center}
\end{table}

\section{Extended Results}\label{appendix:results}

This section presents additional results and an explanation for DVDN's performance on the MPE (Tag) scenario.

\subsection{Level-based foraging}

\begin{figure}[!tb]
    \centering
    \begin{subfigure}{0.32\linewidth}
        \includegraphics[width=\linewidth]{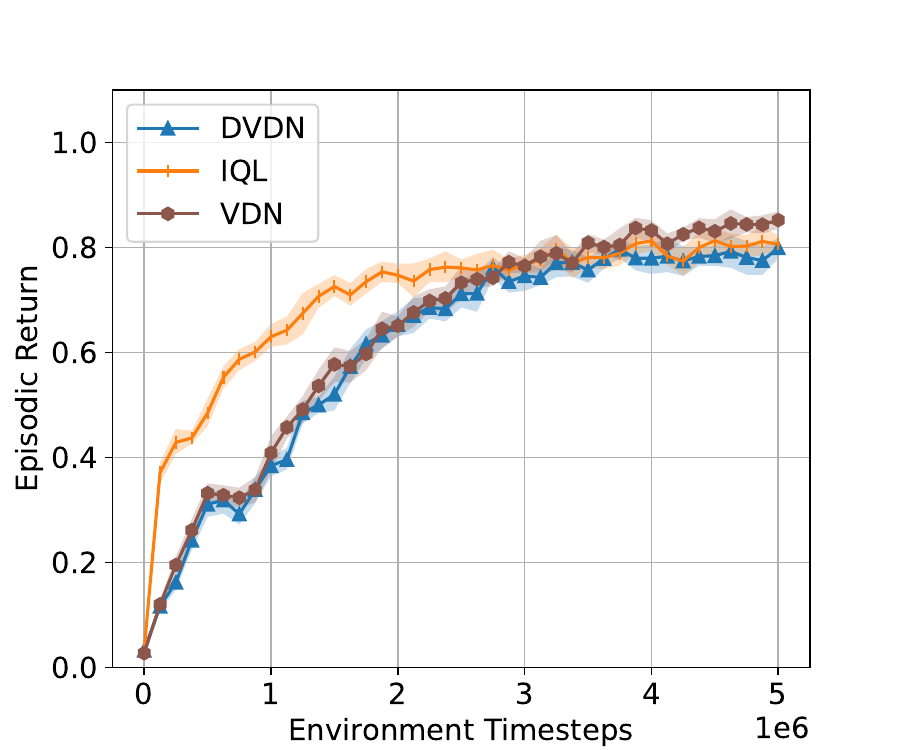}
        \caption{Easy}
    \end{subfigure}
    \begin{subfigure}{0.32\linewidth}
        \includegraphics[width=\linewidth]{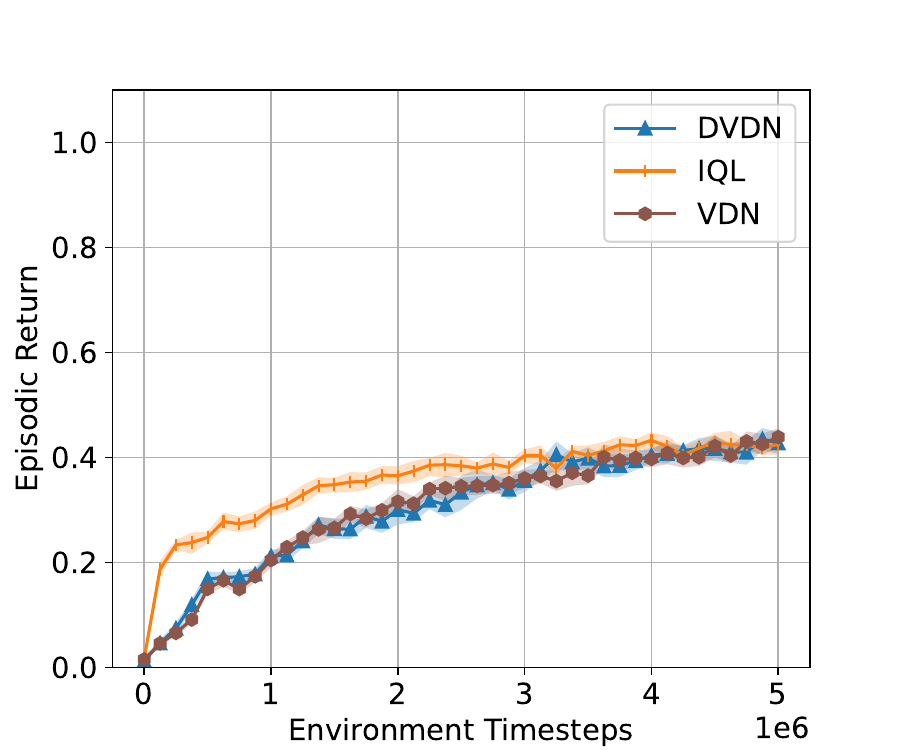}
        \caption{Hard}
    \end{subfigure}
    \caption{The remaining performance plots of the algorithms for heterogeneous agents in the LBF environment, with IQL represented in orange, VDN in chestnut, DVDN in blue. The markers represent the average episodic returns and the shaded area represent the 95\% bootstrap CIs. All algorithms have about the same performance. Particularly, DVDN's learning curve is similar to VDN's.}\label{fig:LBF_ns_test} 
    \end{figure}

Figure~\ref{fig:LBF_ns_test}~(a) and (b) depict the learning curves for DVDN's performance in the Easy and Hard instances of the LBF environment, respectively. The results corroborate those shown in Figure \ref{fig:test}~(a), indicating that all three algorithms have similar performance, and DVDN's learning curve mirrors VDN's learning curve, despite information loss resulting from switching topology dynamics. DVDN  better approximates the performance of VDN. 

\begin{figure}[!tb]
    \centering
    \begin{subfigure}{0.32\linewidth}
        \centering
        \includegraphics[width=\linewidth]{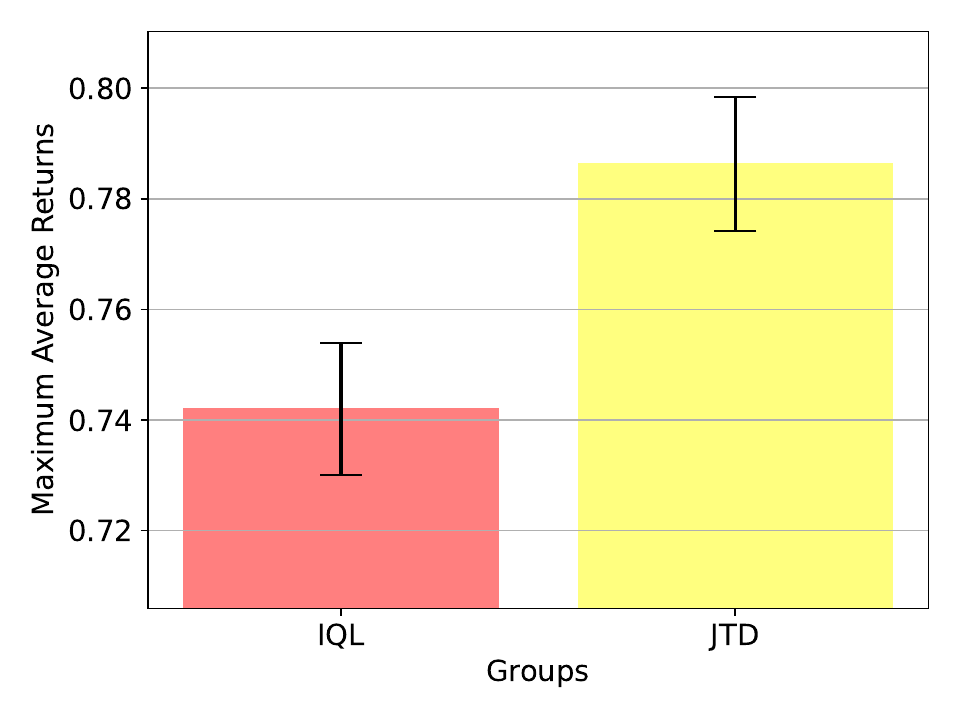}
        \caption{Easy}
    \end{subfigure}
    \begin{subfigure}{0.32\linewidth}
        \centering
        \includegraphics[width=\linewidth]{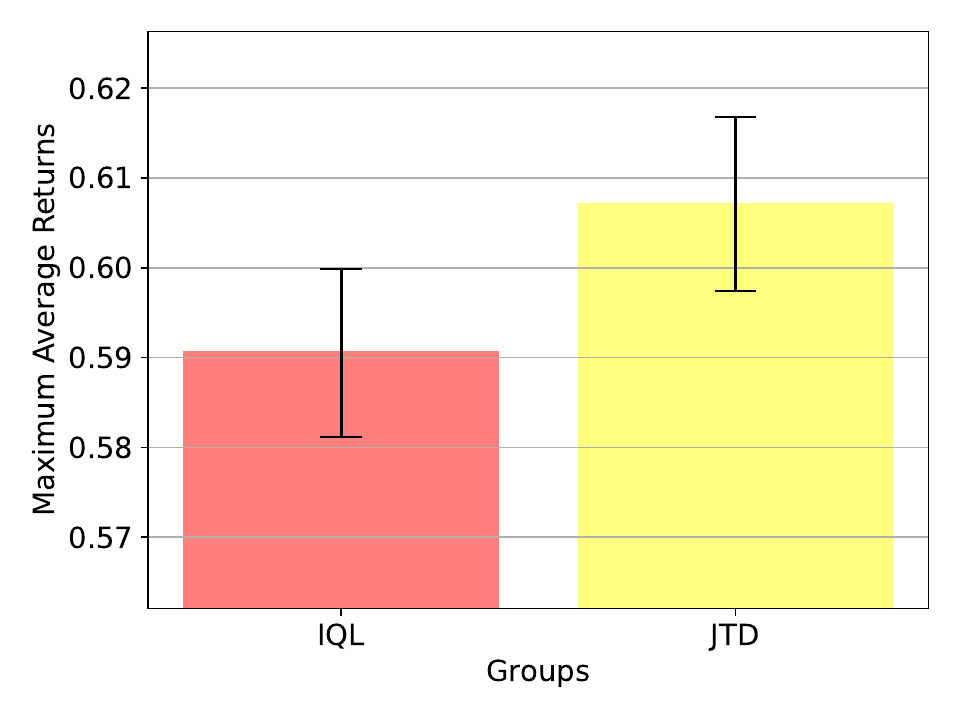}
        \caption{Medium}
    \end{subfigure}
    \begin{subfigure}{0.32\linewidth}
        \centering
        \includegraphics[width=\linewidth]{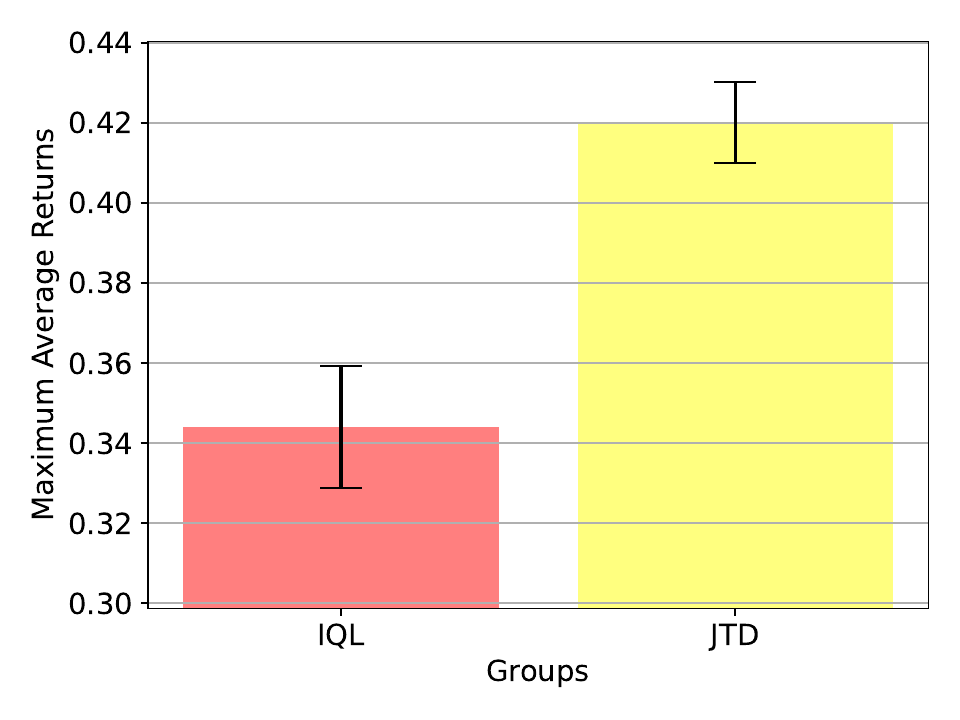}
        \caption{Hard}
    \end{subfigure}
    \caption{Ablation plots for heterogeneous agents in the LBF environment. The IQL (control) group  has no consensus, while JTD group  has joint temporal difference consensus. Notably, JTD consensus leads to significant improvement in results across tasks. }\label{fig:LBF_ns_abls}  
\end{figure}

Figures \ref{fig:LBF_ns_abls}~(a), (b), and (c) illustrate ablation plots for DVDN's key component: joint temporal difference consensus. The inclusion of JTD consensus improves performance across the three tasks (higher is better).

\begin{figure}[!tb]
    \centering
    \begin{subfigure}{0.32\linewidth}
        \includegraphics[width=\linewidth]{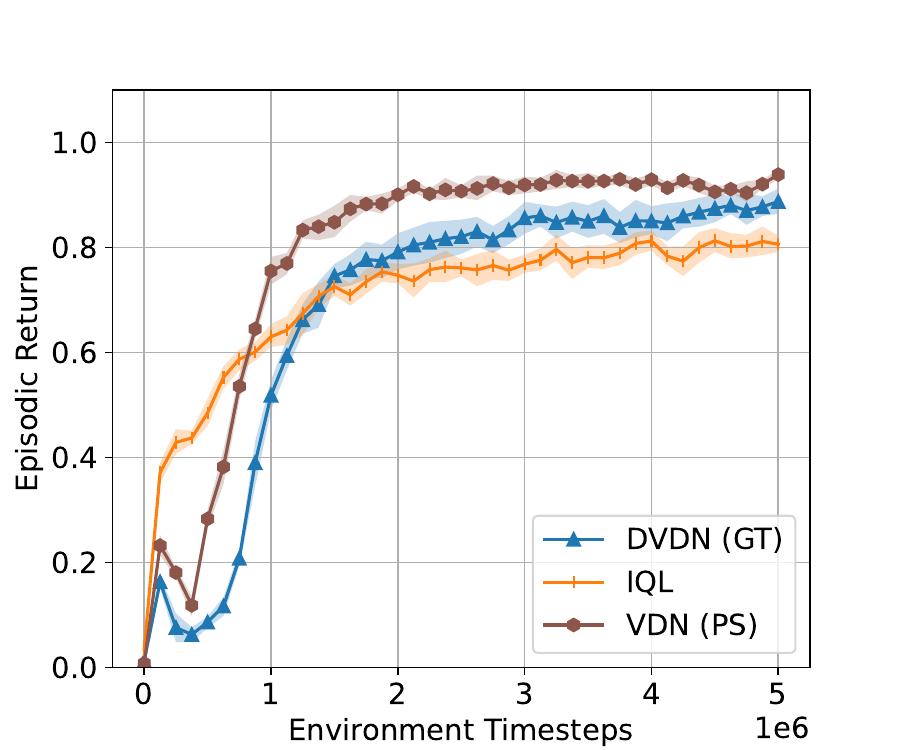}
        \caption{Easy}
    \end{subfigure}
    \begin{subfigure}{0.32\linewidth}
        \includegraphics[width=\linewidth]{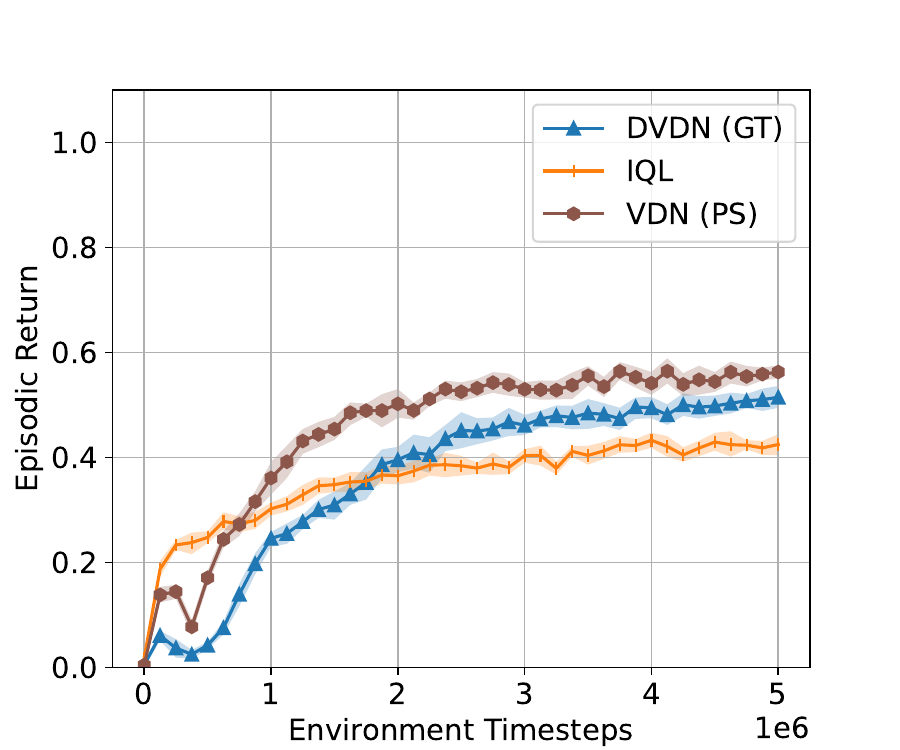}
        \caption{Hard}
    \end{subfigure}
    \caption{The remaining performance plots of the algorithms for homogeneous agents in the  LBF environment, with  IQL represented in orange, VDN (PS) in chestnut, and DVDN (GT) in blue. The markers represent the average episodic returns and the shaded area represents the 95\% bootstrap CIs. All algorithms exhibit comparable performance, with DVDN's learning curve resembling VDN's.}\label{fig:LBF_gt_test}
\end{figure}

\begin{figure}[!tb]
    \centering
    \begin{subfigure}{0.32\linewidth}
        \centering
        \includegraphics[width=\linewidth]{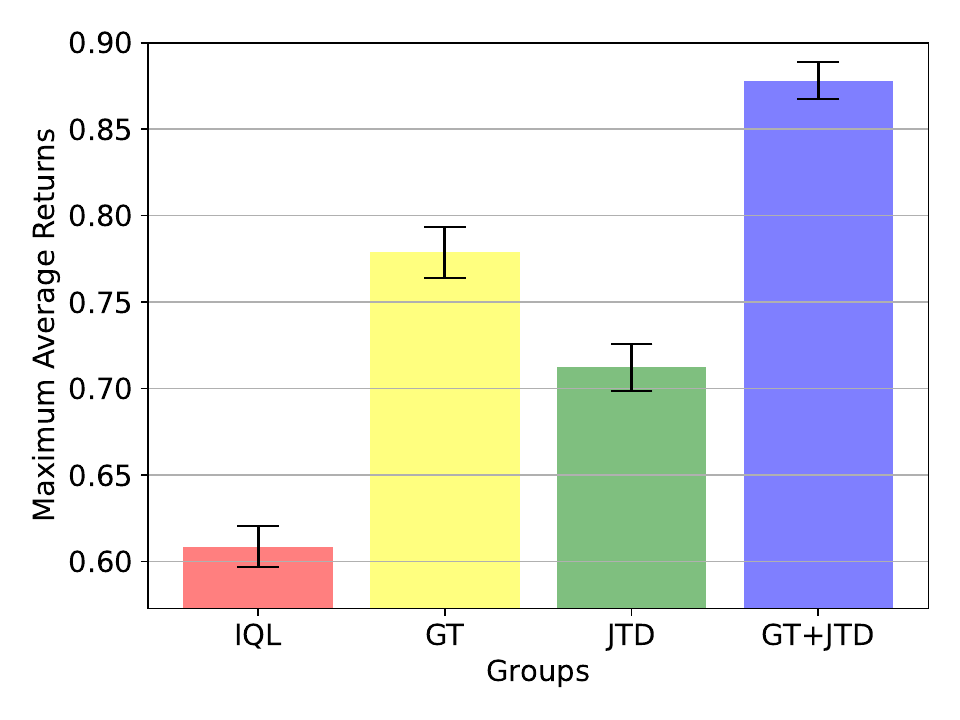}
        \caption{Easy}
    \end{subfigure}
    \begin{subfigure}{0.32\linewidth}
        \centering
        \includegraphics[width=\linewidth]{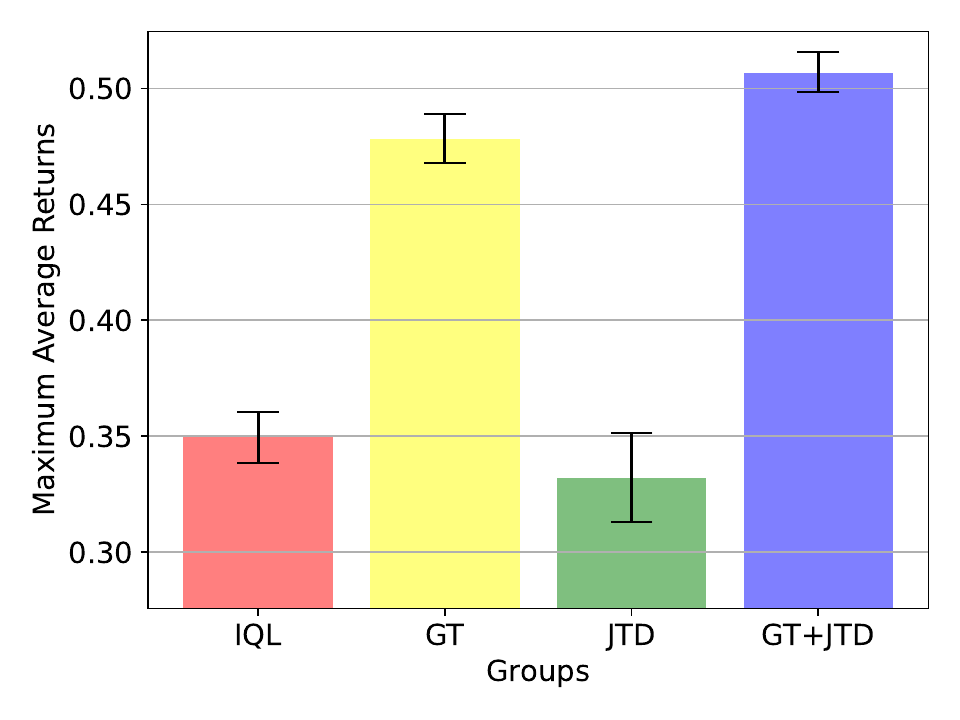}
        \caption{Hard}
    \end{subfigure}
    \caption{Extra ablation plots for homogeneous agents in the LBF environment. The IQL (control) group has no consensus. The GT group performs gradient tracking. The JTD group performs joint temporal difference consensus. The GT+JTD group combines gradient tracking and JTD consensus. In (a) both JTD consensus and gradient tracking independently improve performance, and are better combined. Conversely, in (b) JTD by itself decreases the performance, but when combined with GT it shows superior results. }\label{fig:LBF_gt_abls}
\end{figure}

Figure~\ref{fig:LBF_gt_test} (a) and (b) show the learning curves for DVDN (GT) in the LBF environment for Easy and Hard instances, respectively. These results support previous findings in Figure~\ref{fig:test} (a), where DVDN (GT)'s learning curve approximates to VDN (PS)'s. In the homogeneous agents setting, parameter sharing and gradient tracking significantly improve the performance of both algorithms. Figure~\ref{fig:LBF_gt_abls} (a) and (b) provide ablation plots for DVDN (GT)'s two main components: joint temporal difference (JTD) and gradient tracking (GT). Interestingly, JTD degrades performance individually in the hard task (Figure~\ref{fig:LBF_gt_abls} (b)), but combining it with GT yields better results than using GT alone. As the task difficulty increases, the benefit of JTD diminishes. We speculate that for hard tasks, a near optimal policy involves independent behavior, such as walking to the nearest fruit and attempting a load action. This finding is supported by previous work~\citep{papoudakis_2021}.

\subsection{Multi -particle environment}

\begin{figure}[!tb]
    \centering
    \begin{subfigure}{0.32\linewidth}
        \includegraphics[width=\linewidth]{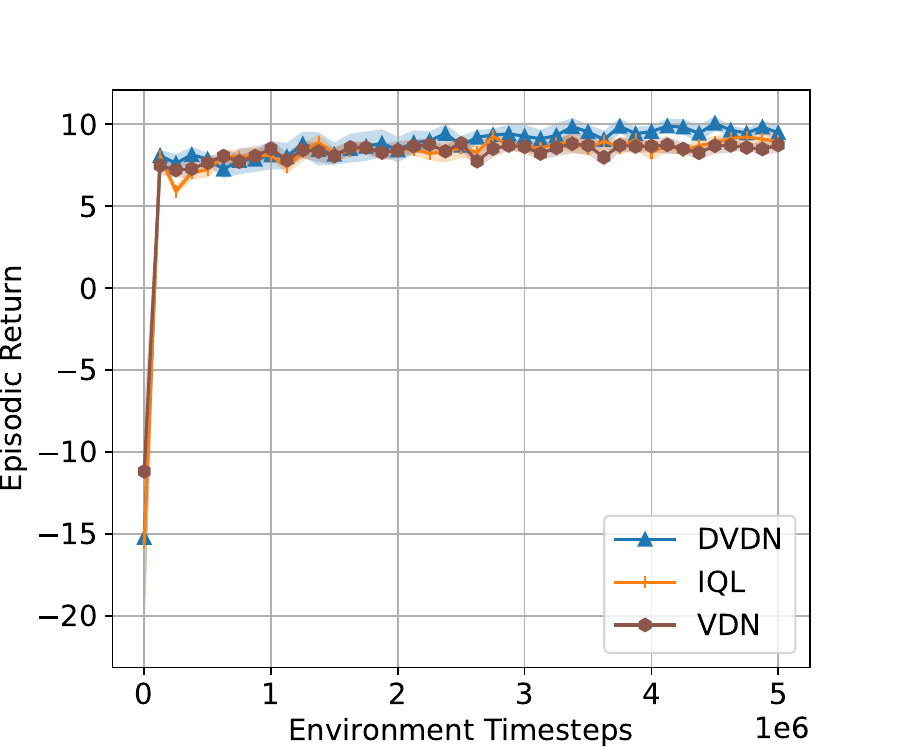}
        \caption{Adversary}
    \end{subfigure}
    \begin{subfigure}{0.32\linewidth}
        \includegraphics[width=\linewidth]{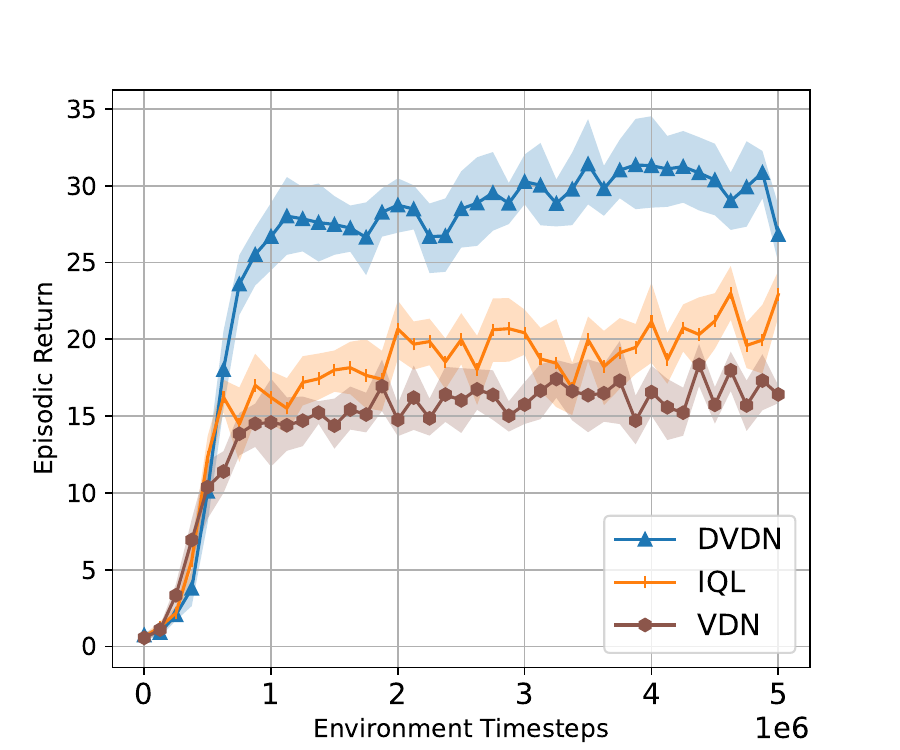}
        \caption{Tag}
    \end{subfigure}
    \caption{The remaining performance plots of the algorithms for heterogeneous agents in the MPE environment,  with IQL represented in orange, VDN in chestnut, and  DVDN in blue. The markers represent the average episodic returns and the shaded area represents the 95\% bootstrap CIs. For scenario (a)  algorithms have about the same sample-efficiency making it hard to discriminate between different approaches.  However, for scenario (b), DVDN outperforms other approaches.}\label{fig:MPE_ns_test}  
    \end{figure}

Figure~\ref{fig:MPE_ns_test} (a) and (b) show the learning curves for DVDN's performance, heterogeneous setting, for the individual scenarios belonging to the MPE environment: Adversary, and Tag tasks respectively.  For Adversary (and Spread) that have dense rewards, algorithms have similar sample-efficiency. However, the best policies for Adversary are generated by DVDN, while the best policies for Spread are  generated by IQL. For those scenarios it is hard to assess differences in performances between DVDN and VDN. For the Tag scenario, we see that both DVDN and VDN present fast performance growth in the first million timesteps and have their highest performing average episodic returns close to timestep four million. The policies generated however, are very different for reasons we elaborate in a qualitative analysis in Appendix~\ref{appendix:qualitative_analysis}. Figure~\ref{fig:MPE_ns_abls} (a), (b)  and (c) show the ablation plots for DVDN's main component: joint temporal difference consensus. JTD consensus improves the performance for the three tasks (higher better). 

\begin{figure}[!tb]
    \centering
    \begin{subfigure}{0.32\linewidth}
        \centering
        \includegraphics[width=\linewidth]{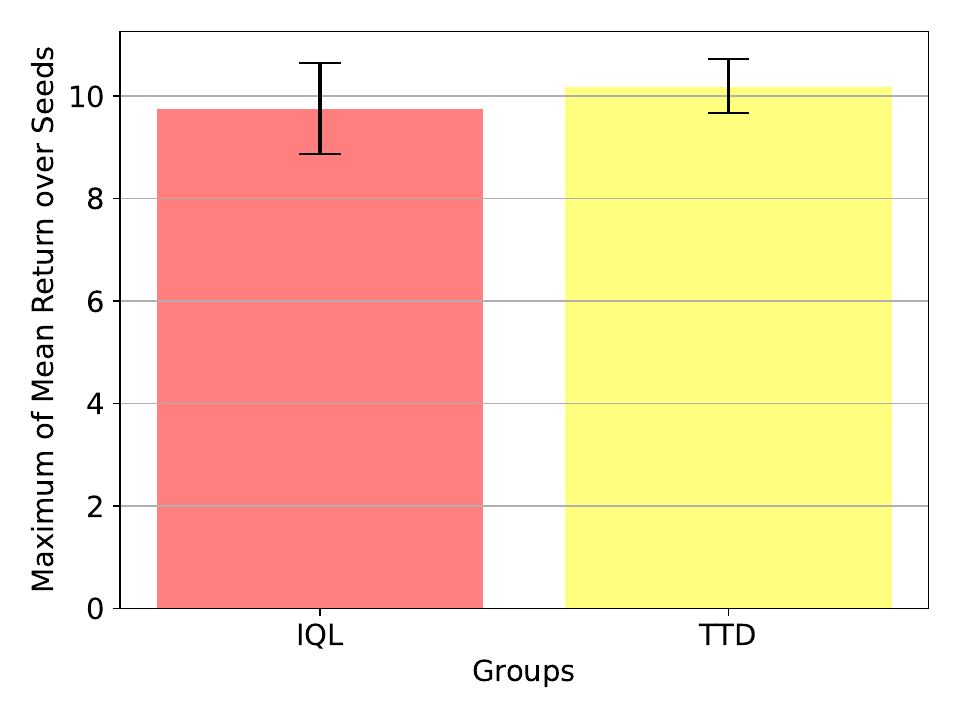}
        \caption{Adversary}
    \end{subfigure}
    \begin{subfigure}{0.32\linewidth}
        \centering
        \includegraphics[width=\linewidth]{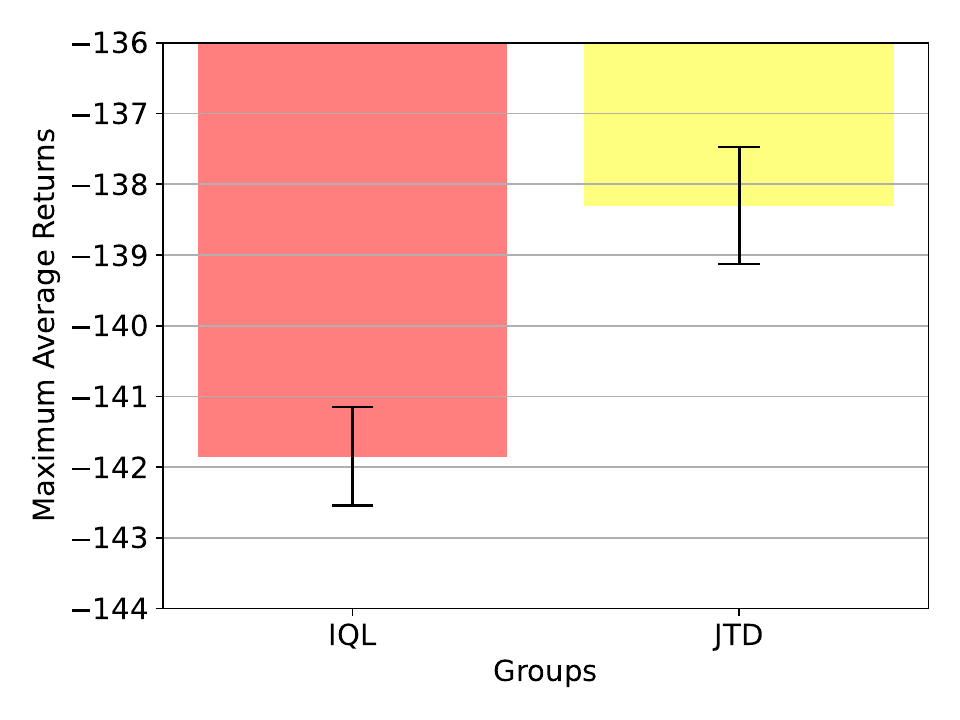}
        \caption{Spread}
    \end{subfigure}
    \begin{subfigure}{0.32\linewidth}
        \centering
        \includegraphics[width=\linewidth]{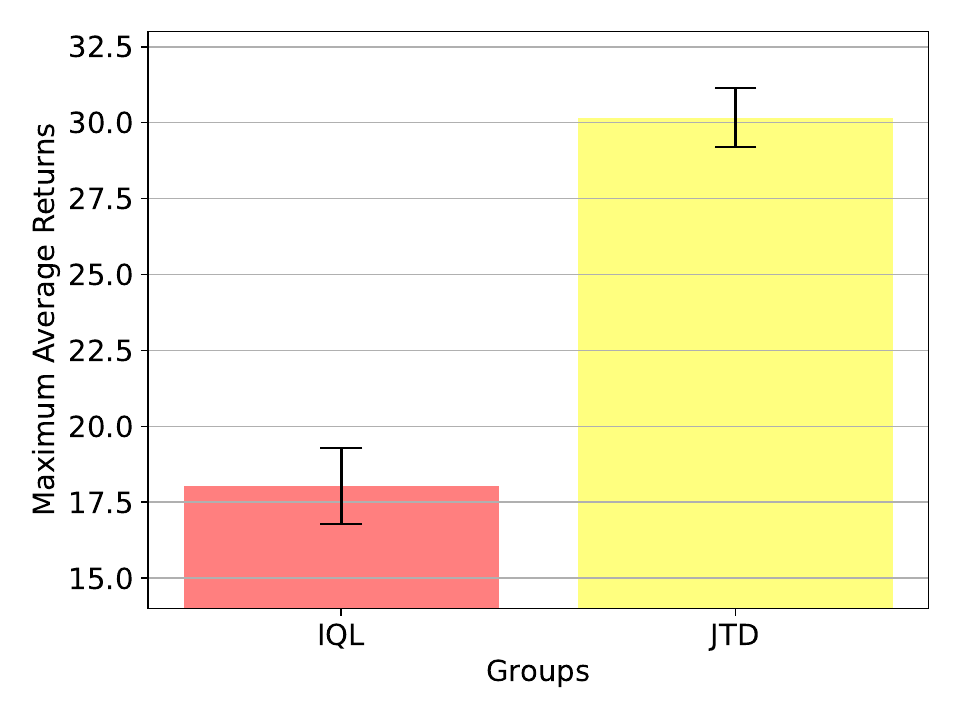}
        \caption{Tag}
    \end{subfigure}
    \caption{Ablation plots for heterogeneous agents in the MPE environment. The IQL (control) group has no consensus, while JTD group has joint temporal difference consensus. JTD consensus leads to significant improvement in results for tasks (b) and (c).}\label{fig:MPE_ns_abls}
\end{figure}

\begin{figure}[!tb]
    \begin{subfigure}{0.32\linewidth}
        \includegraphics[width=\linewidth]{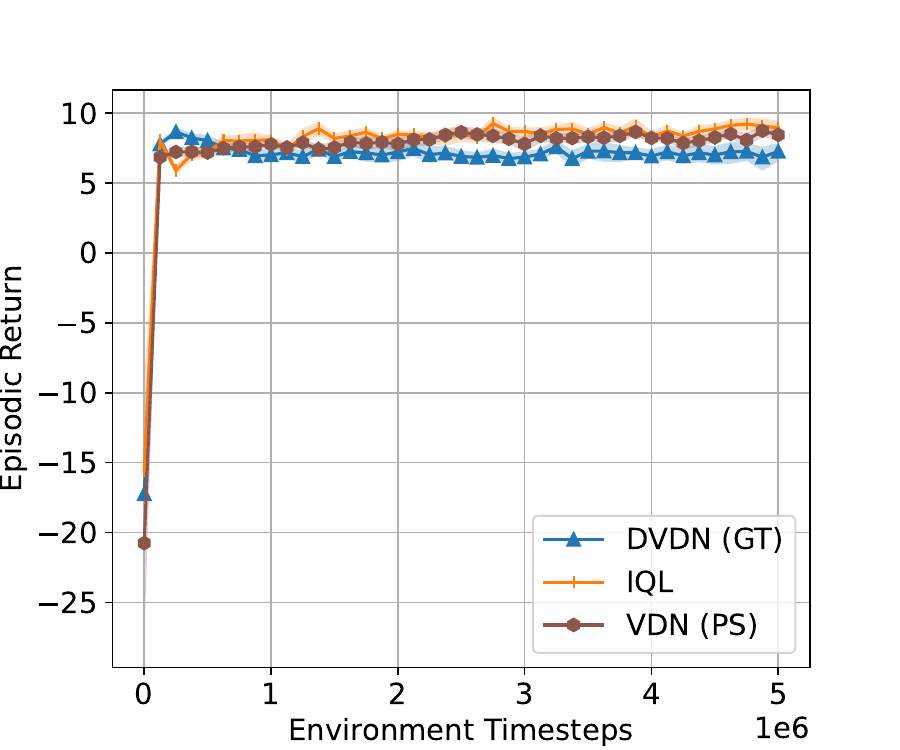}
        \caption{Adversary}
    \end{subfigure}
    \begin{subfigure}{0.32\linewidth}
        \includegraphics[width=\linewidth]{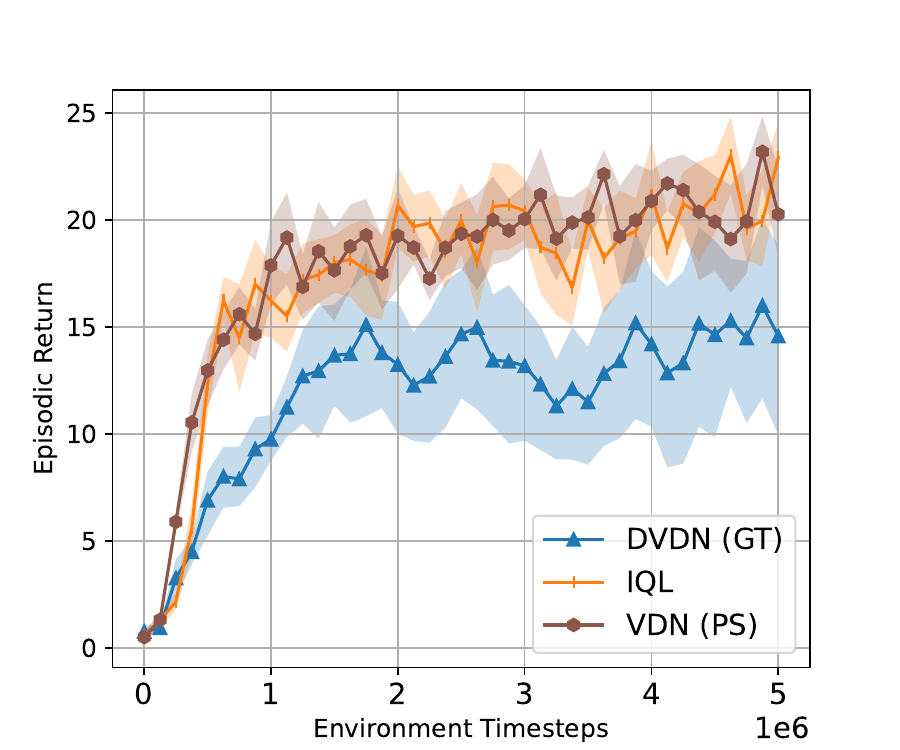}
        \caption{Tag}
    \end{subfigure}
    \caption{The remaining performance plots of the algorithms for homogeneous agents in the  MPE environment, with IQL in orange, VDN (PS) in chestnut  and DVDN (GT) in blue. The markers represent the average episodic returns and the shaded area represent the 95\% bootstrap CIs. DVDN (GT) underperforms other approaches. We expand the qualitative reasons for the under performance of DVDN (GT) in (Appendix \ref{appendix:qualitative_analysis}).}\label{fig:MPE_gt_test}
\end{figure}

Figure~\ref{fig:MPE_gt_test} (a) and (b)  show the learning curves for DVDN (GT) in the MPE environment for Adversary and Tag scenarios, respectively. Similar to the heterogeneous setting, sample efficiency are approximately the same for all algorithms in the Adversary task, making it difficult to distinguish performances between algorithms. In scenario (a), IQL outperforms other algorithms, while in the Spread scenario, DVDN (GT) achieves the best results. However, in the Tag scenario, DVDN (GT) underperforms other approaches. Qualitative analysis (Appendix~\ref{appendix:qualitative_analysis}) reveals that performance comparable to other algorithms was achieved during hyperparameter tuning. Suggesting a limitation in  our (and previous work) experimentation protocol rather than an intrinsic flaw in our design.  Additionally, the 95\% bootstrap CI is wide, indicating that a few of the 10 independent runs generated higher-quality policies.

\begin{figure}[!tb]
    \centering
    \begin{subfigure}{0.32\linewidth}
        \centering
        \includegraphics[width=\linewidth]{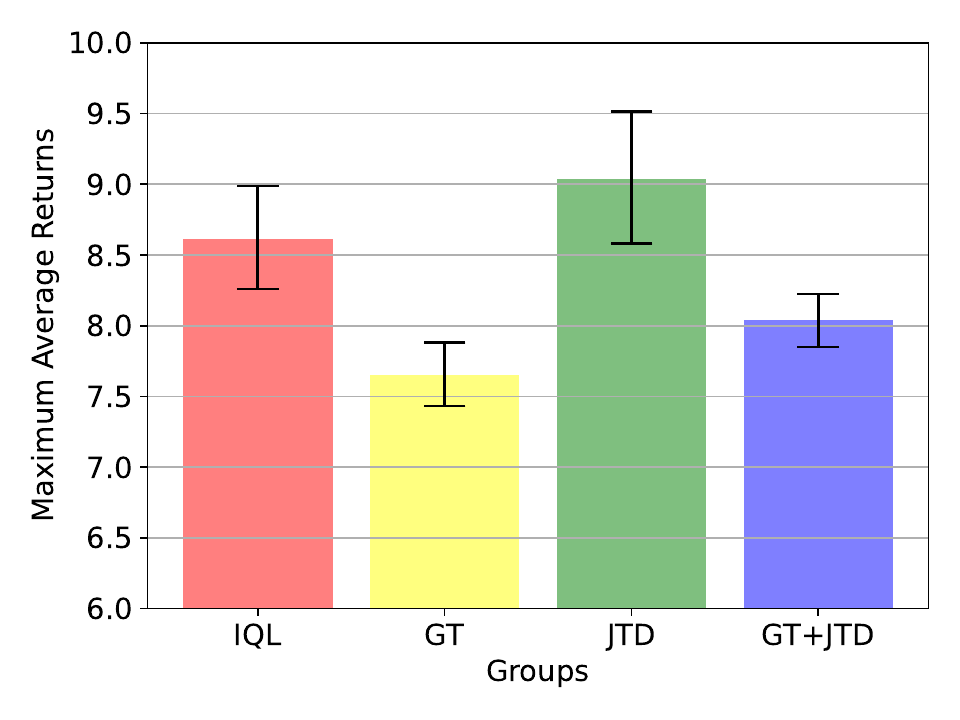}
        \caption{Adversary}
    \end{subfigure}
    \begin{subfigure}{0.32\linewidth}
        \centering
        \includegraphics[width=\linewidth]{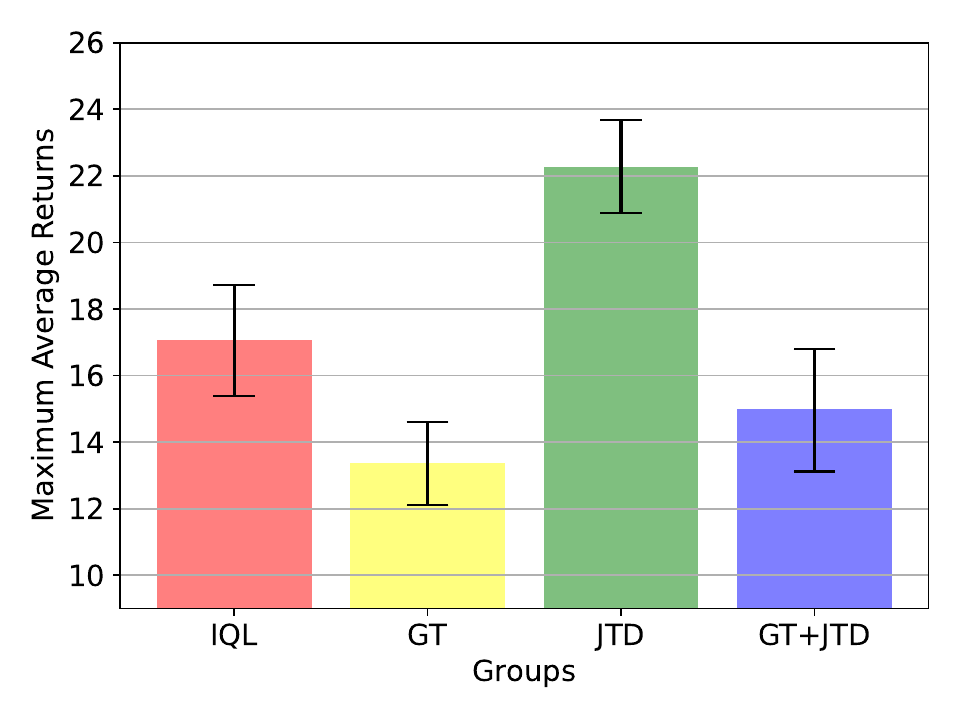}
        \caption{Tag}
    \end{subfigure}
    \caption{Extra ablation plots for the homogeneous in the MPE environment. The IQL (control) group has no consensus. The GT group performs gradient tracking. The JTD group performs joint temporal difference consensus. The GT+JTD group combines gradient tracking and JTD consensus. In (a) and (b) JTD consensus improves performance, while the groups GT and GT+JTD underperforms IQL. }\label{fig:MPE_gt_abls}
\end{figure}

Figure~\ref{fig:MPE_gt_abls} (a) and (b) display ablation plots for DVDN (GT)'s two main components: joint temporal difference consensus (JTD) and gradient tracking (GT). For the Tag scenario (Figure~\ref{fig:MPE_gt_abls} (b)), JTD improves performance, while the improvement is marginal for the Adversary scenario (Figure~\ref{fig:MPE_gt_abls} (a)). Conversely, GT degrades performance compared to IQL in the Adversary and Tag scenarios (Figure~\ref{fig:MPE_gt_abls} (a) and (b)). Combining GT and JTD consensus (group GT+JTD) results in decreased performance compared to applying JTD consensus alone.

\subsection{MARBLER}

\begin{figure}[!tb]
    \begin{subfigure}{0.32\linewidth}
        \includegraphics[width=\linewidth]{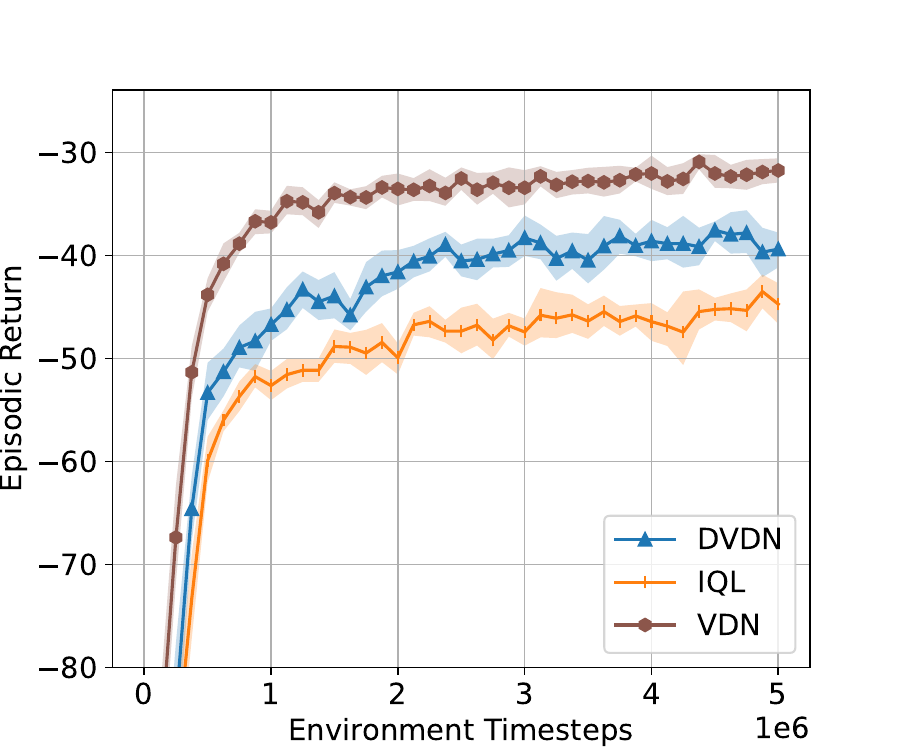}
        \caption{Arctic}
    \end{subfigure}
    \begin{subfigure}{0.32\linewidth}
        \includegraphics[width=\linewidth]{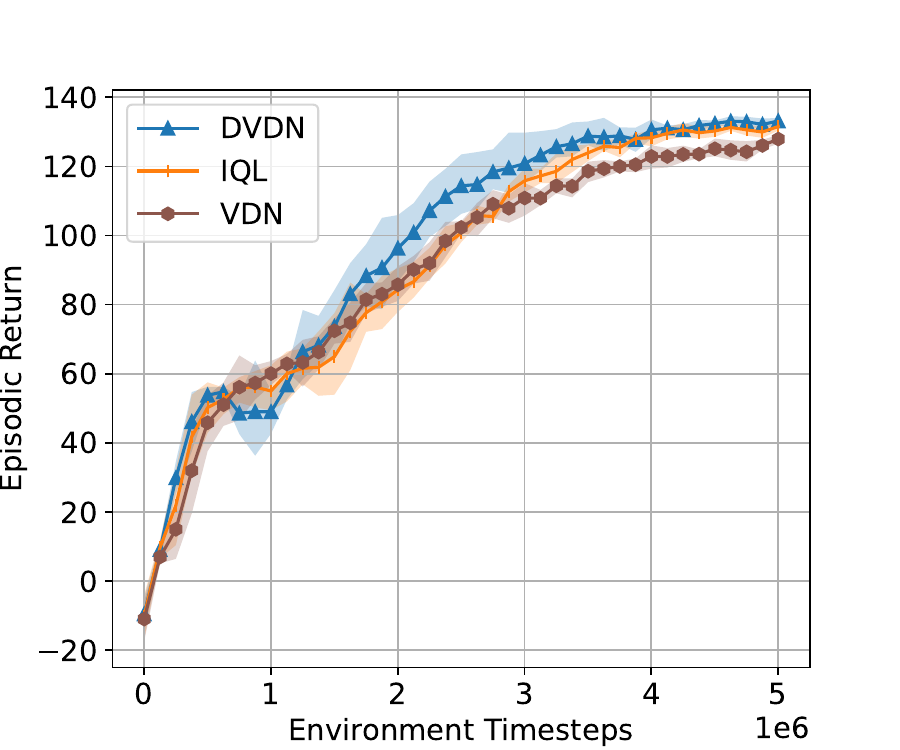}
        \caption{PCP}
    \end{subfigure}
    \begin{subfigure}{0.32\linewidth}
        \includegraphics[width=\linewidth]{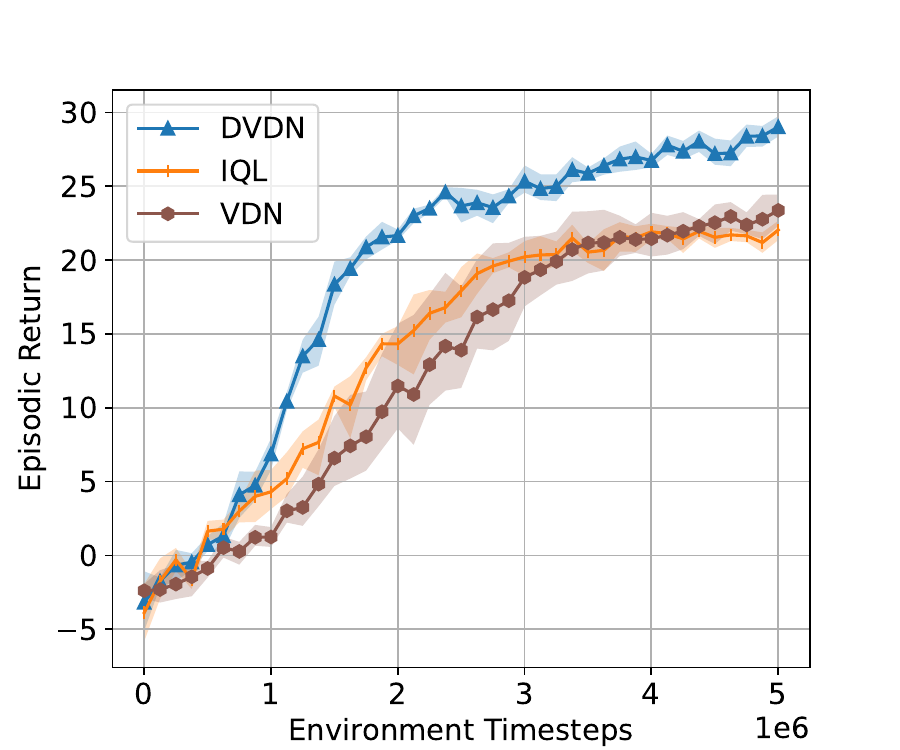}
        \caption{Warehouse}
    \end{subfigure}
    \caption{The remaining performance plots of the algorithms for heterogeneous agents in the MARBLER environment, with IQL represented in orange, VDN in chestnut, and DVDN in blue. The markers represent the average episodic returns and the shaded area represent the 95\% bootstrap CIs. For scenario (a), (b), and (c) DVDN and VDN present similar behavior during training. DVDN outperforms with margin IQL for scenarios (a) and (c), and VDN for scenarios (b) and (c).}\label{fig:MARBLER_ns_test} 
    \end{figure}

Figure~\ref{fig:MARBLER_ns_test} (a), (b), and (c) show the learning curves for DVDN's performance in the heterogeneous agents setting, for the remaining scenarios belonging to the MARBLER environment: Arctic, PCP and Warehouse tasks respectively. DVDN and VDN have similar learning curves for the three scenarios, but DVDN  outperforms VDN in scenarios (b) and (c). In scenario (c), IQL outperforms VDN, but both algorithms are still surpassed by DVDN. 

\begin{figure}[!tb]
    \centering
    \begin{subfigure}{0.24\linewidth}
        \centering
        \includegraphics[width=\linewidth]{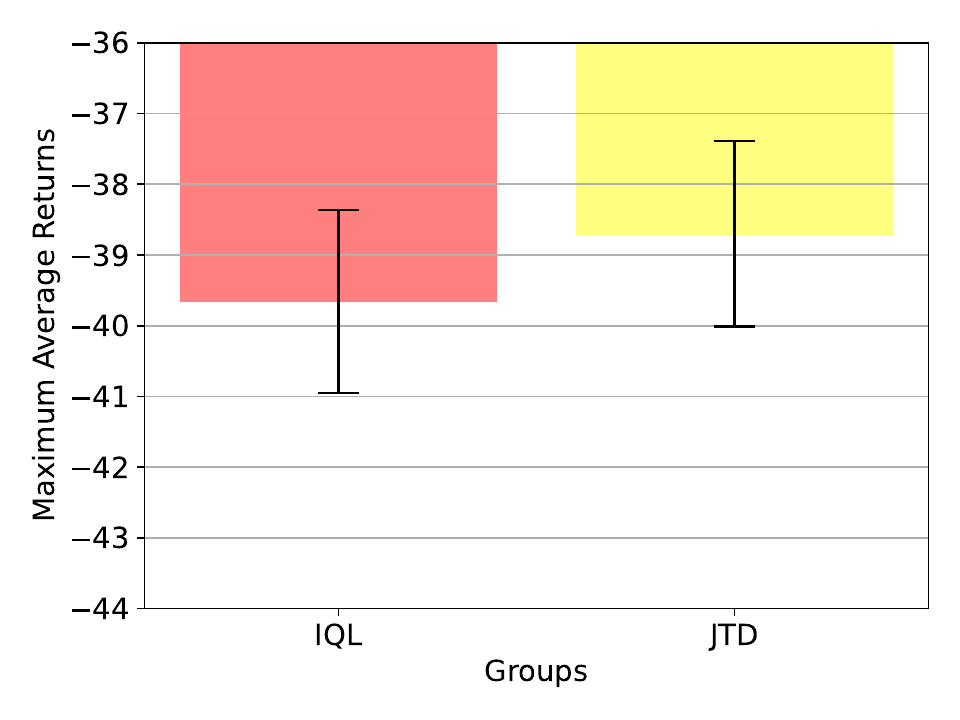}
        \caption{Arctic}
    \end{subfigure}
    \begin{subfigure}{0.24\linewidth}
        \centering
        \includegraphics[width=\linewidth]{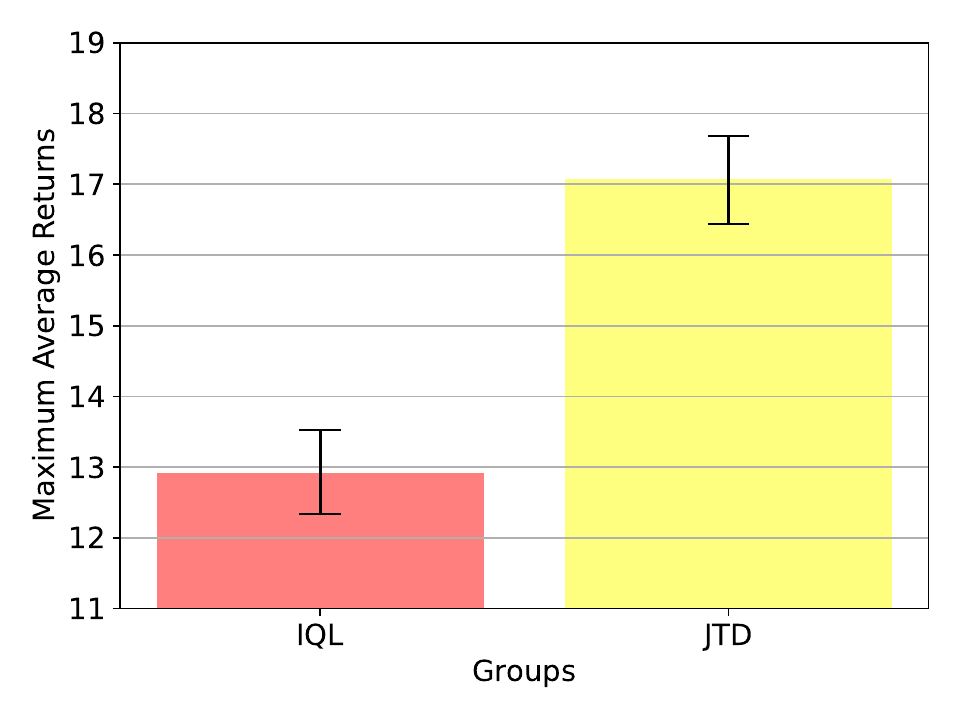}
        \caption{Material}
    \end{subfigure} 
    \begin{subfigure}{0.24\linewidth}
        \centering
        \includegraphics[width=\linewidth]{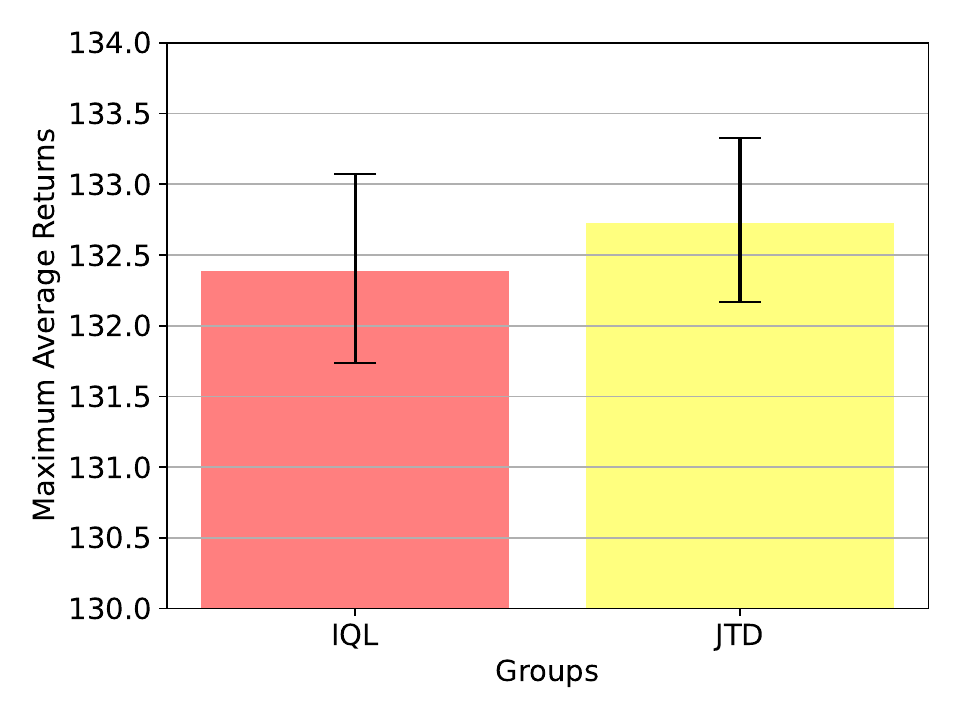}
        \caption{PCP}
    \end{subfigure}
    \begin{subfigure}{0.24\linewidth}
        \centering
        \includegraphics[width=\linewidth]{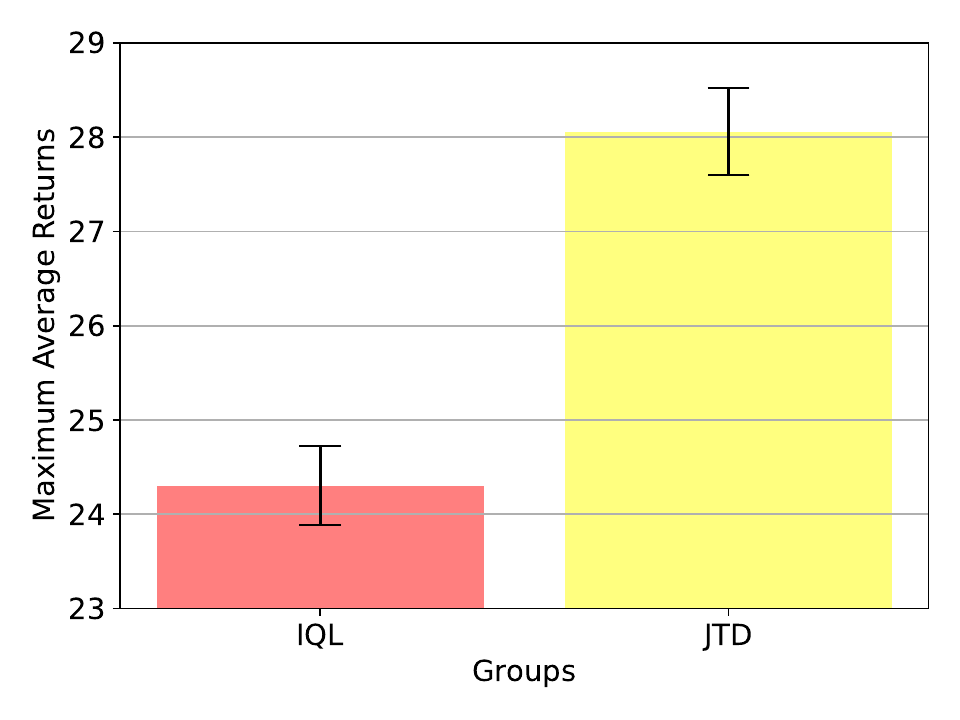}
        \caption{Warehouse}
    \end{subfigure} 
    \caption{Ablation plots for heterogeneous agents in the MARBLER environment. The IQL group has no consensus, while JTD group (control group) has joint temporal difference consensus. For scenarios (b) and (d), the JTD consensus improves results significantly. Whereas, for scenarios (a) and (c), JTD consensus improves results marginally.}\label{fig:MARBLER_ns_abls}
\end{figure}

Figure~\ref{fig:MARBLER_ns_abls} (a), (b), (c) and (d) show the ablation plots for DVDN's main component, joint temporal difference consensus, for the scenarios Arctic,  Material, PCP and Warehouse. JTD consensus enhances performance for the scenarios (b) and (d), with minor improvements in scenarios (a) and (c). 

\begin{figure}[!tb]
    \centering
    \begin{subfigure}{0.33\linewidth}
       \centering
        \includegraphics[width=\linewidth]{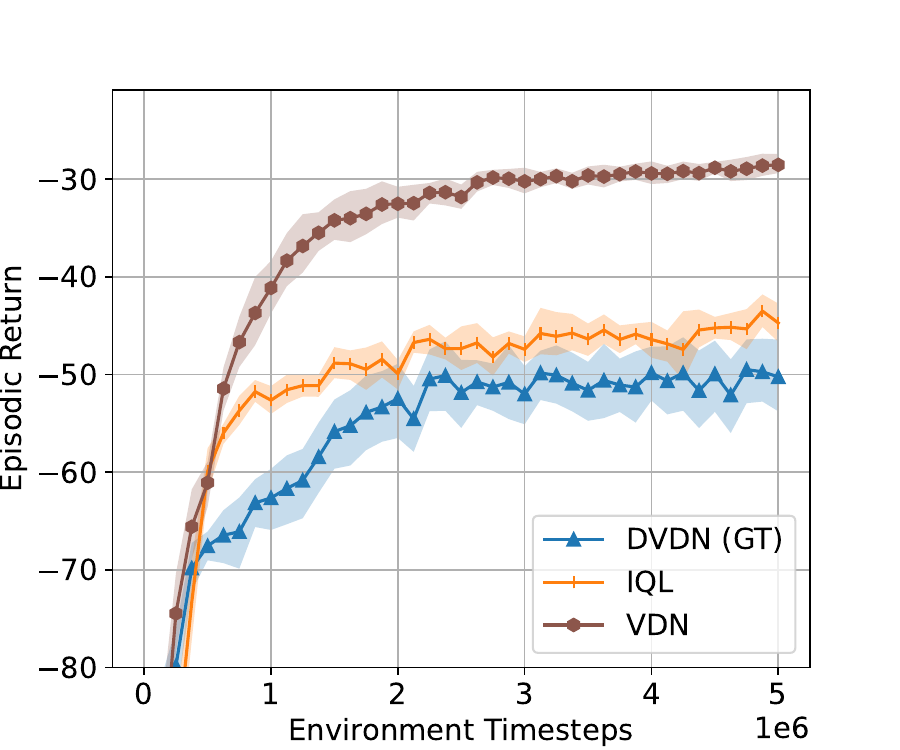}
        \caption{Arctic}
    \end{subfigure}
    \begin{subfigure}{0.32\linewidth}
        \includegraphics[width=\linewidth]{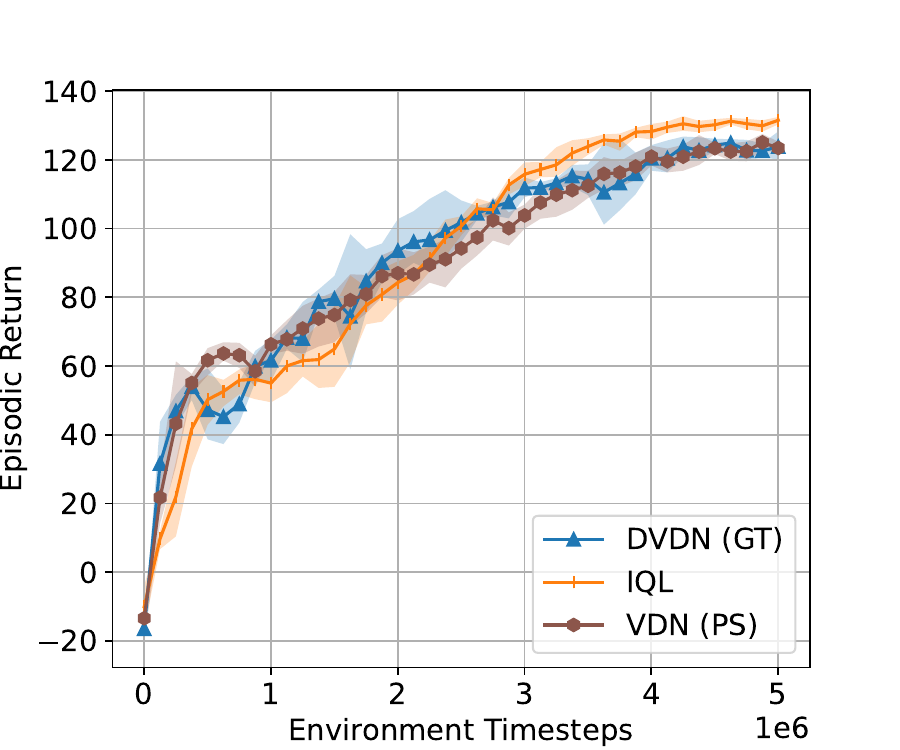}
        \caption{PCP}
    \end{subfigure}
    \begin{subfigure}{0.32\linewidth}
        \includegraphics[width=\linewidth]{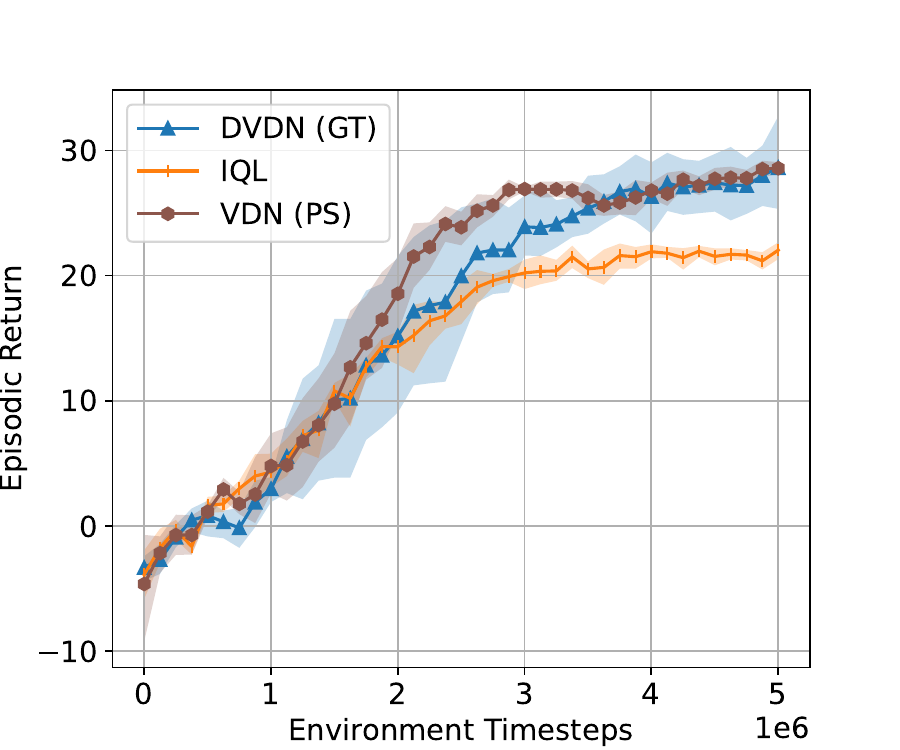}
        \caption{Warehouse}
    \end{subfigure}
    \caption{The remaining performance of the algorithms for homogeneous agents in the  MARBLER environment, with  IQL in orange, VDN (PS) in chestnut  and DVDN (GT) in blue. The markers represent the average episodic returns and the shaded area represent the 95\% bootstrap CIs. In scenario (a) and (b), DVDN (GT) is outperformed by IQL, while in scenario (c) DVDN (GT) outperforms IQL.}\label{fig:MARBLER_gt_test}
\end{figure}

 Figure~\ref{fig:MARBLER_gt_test} (a), (b), and (c) show the learning curves for DVDN (GT) performance in the MARBLER environment for Arctic, PCP and Warehouse scenarios respectively. In scenario (a), DVDN (GT)'s learning curve is similar to IQL's, but it underperforms in this highly heterogeneous agents scenario. Gradient tracking assumes that all agents follow a  similar policy, but in this assumption is mostly violated in Arctic as there are agents with different roles. Whereas VDN (PS) show a modest improvement moving from the heterogeneous setting, DVDN (GT) shows a degradation in performance. We hypothesize that differences from VDN's loss function and DVDN (GT)'s loss function~\eqref{eqn:GT_unconstrained} plays a key role in this result. However, in scenarios (a) and (b), DVDN (GT)'s learning curve is similar to VDN (PS)'s.

\begin{figure}[!tb]
    \begin{subfigure}{0.33\linewidth}
         \centering
        \includegraphics[width=\linewidth]{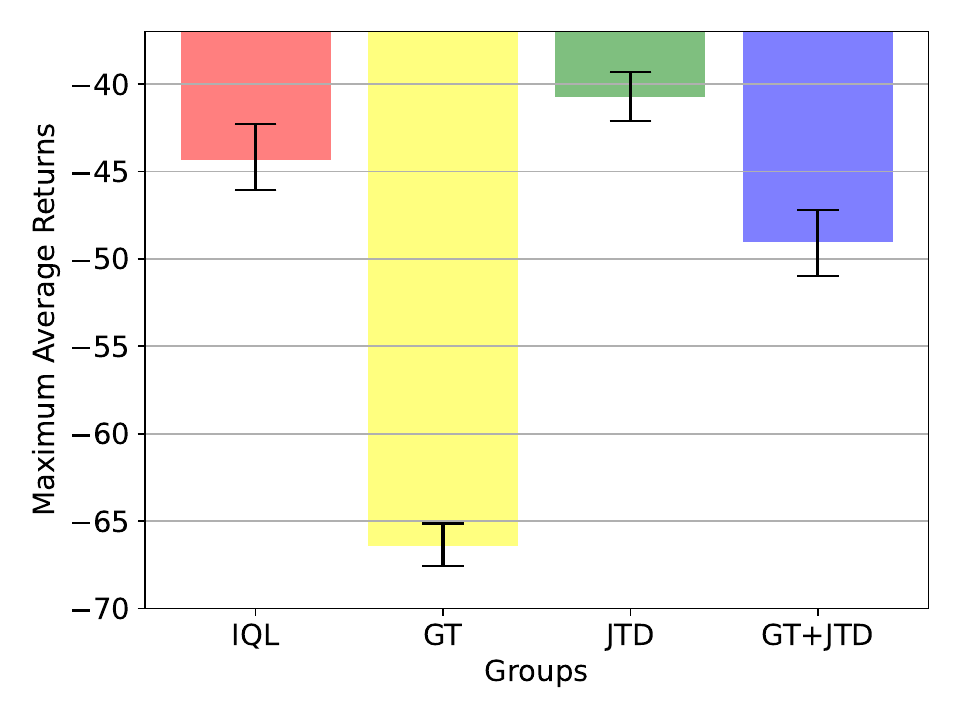}
        \caption{Arctic}
    \end{subfigure}
    \begin{subfigure}{0.33\linewidth}
         \centering
        \includegraphics[width=\linewidth]{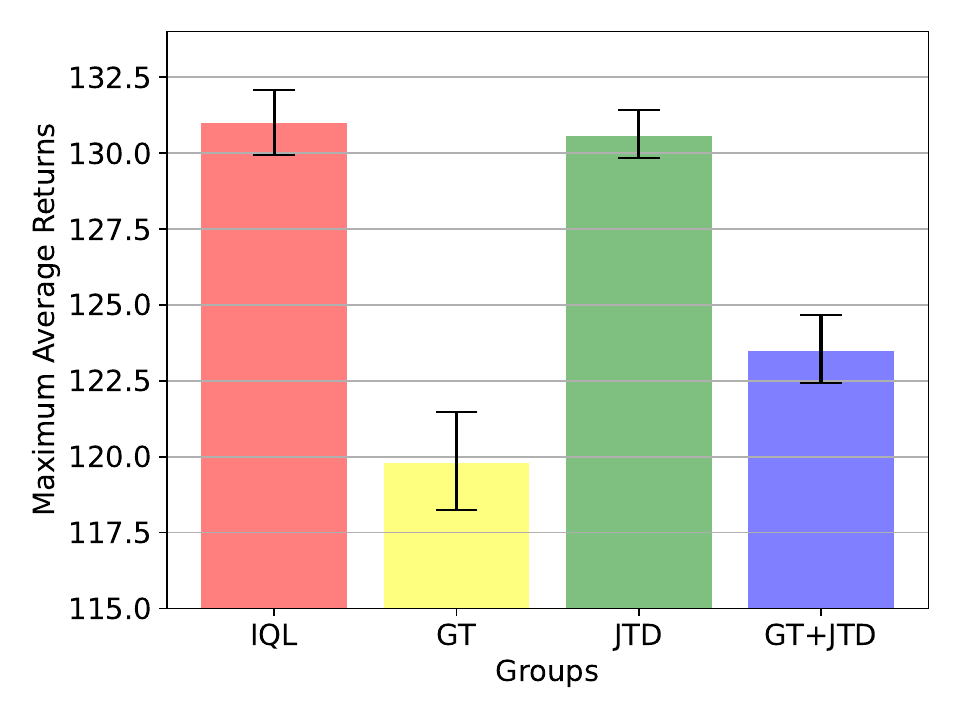}
        \caption{PCP}
    \end{subfigure}
    \begin{subfigure}{0.33\linewidth}
        \centering
        \includegraphics[width=\linewidth]{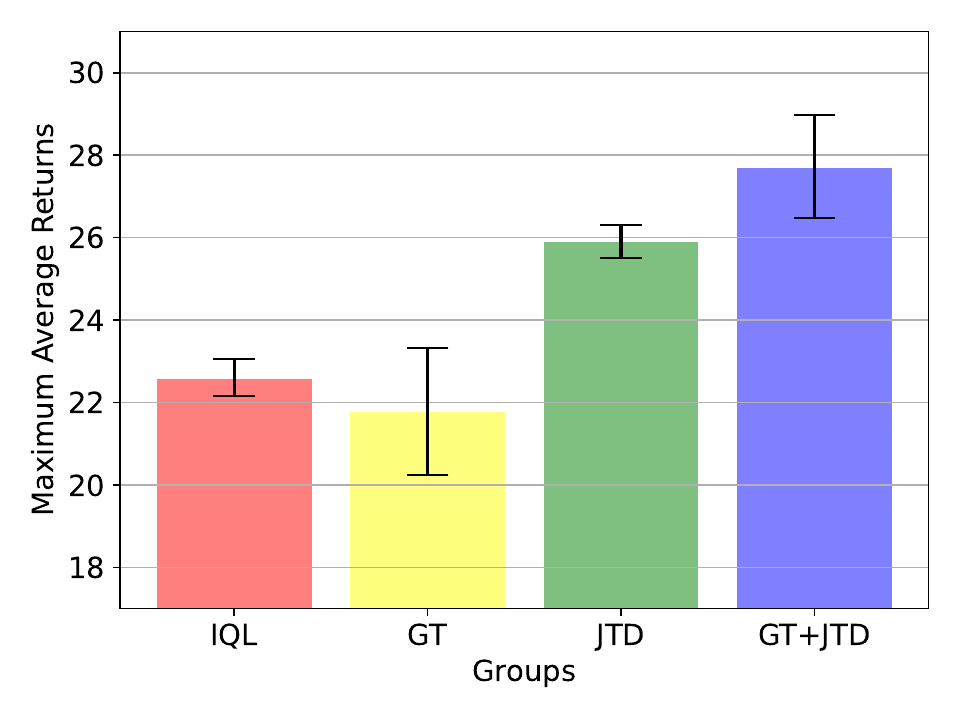 }
        \caption{Warehouse}
    \end{subfigure}
    \caption{Extra ablation plots for the homogeneous in the MARBLER environment. The IQL (control) group has no consensus. The GT group performs gradient tracking. The JTD group performs joint temporal difference consensus. The GT+JTD group  combines gradient tracking and JTD consensus. In scenario (c), JTD consensus and GT combined improve the performance. Conversely, in (a) and (b),  GT+JTD the performance degrades. }\label{fig:MARBLER_gt_abls}
\end{figure}

 Figure~\ref{fig:MARBLER_gt_abls} (a), (b), and (c) show ablation plots for DVDNs (GT)'s two main components: joint temporal difference consensus and gradient tracking. For the three scenarios GT degrades performance The net effect of combining both GT and JTD consensus (group GT+JTD) is beneficial in scenario (c).

\subsection{Qualitative Analysis}\label{appendix:qualitative_analysis}

In this section, we elaborate on the high performance for DVDN as observed in Table~\ref{tab:max} and Figure~\ref{fig:MPE_ns_test}. Despite being designed to emulate VDN, DVDN surpasses both baselines by a substantial margin in Figure \ref{fig:MPE_ns_test}, which is unexpected.  A qualitative analysis of the optimal joint policy for VDN reveals that shows two agents assume the blocker role, performing  random walk, which is effective in  restraining the prey's movement while a third agent acts as the pursuer, chasing the prey and relying on the blockers to corner it. The behavior policy of DVDN is different: Two agents act as runners, while the third serves the pursuer. The runners agents flee from the grid,  leaving its regular bounds. The prey adopts a passive behavior, allowing the pursuer to bump into it. These discrepancies rise two questions: {\em How is this possible?} And {\em why does it work?}

As to how two algorithms that are supposed to generate the same kind of behavior, instead generate such distinct behaviors, the answer is exploration. Although, DVDN's learning curve is supposed to approach VDN's during random exploration the learning paths may diverge wildly. Moreover, controlled state exploration is one factor separating the central agent from the centralized learners. DVDN's  estimation of JTD may lead to a greater variety of behavior due to information loss. In this case, those exploratory behaviors turned out to be useful. As to why does this apparently odd behavior works. It works because the prey agent was trained using deep reinforcement learning \cite[DDPG,][]{lillicrap_2016} and its policy is fixed, making the prey agent part of the environment from the perspective of the training predator agents. During random state space exploration, predators generate new unseen states by the prey. Subsequently, the prey's policy shows a poor generalization, it cannot reason about behaviors of the predators. This effect is called adversarial training, where the adversaries are the predators.~\citet{gleave_2020} found that adversaries win by generating states that are adversarial, \ie, that exploit  the pretrained victim's policy, not particularly by behaving strategically.   

In Figure~\ref{fig:MPE_gt_test} (b),  DVDN (GT) underperforms  the baselines. Figure~\ref{fig:MPE_gt_abls} (b)  reveals that GT has a detrimental impact on performance when compared to IQL group. Nevertheless, this configuration was the best performing during hyperparameter grid search. In fact, if we replace the ten randomly seeded test evaluations by the three sequential seeded hyperparameter grid search trials, performance improves drastically. DVDN (GT) matches baselines' performance (see  Fig.~\ref{fig:replacement} and Table~\ref{tab:replacement}).   This finding suggests a limitation~\citep{papoudakis_2021}'s hyperparameter search protocol.  Three trials may not be sufficient to assess experiments with high variability~\citep{henderson_2018}, such as predator-and-prey.

\begin{table}[!tb]
    \begin{minipage}[b]{0.65\linewidth}
        \centering
        \includegraphics[width=\linewidth]{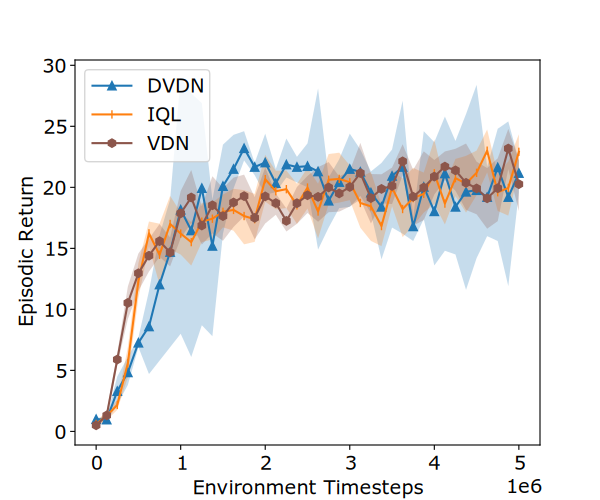}
        \captionof{figure}{Performance for predator-and-prey task, DVDN (GT) results on hyperparameter grid search trials.}\label{fig:replacement}
    \end{minipage}
\begin{minipage}[b]{0.3\linewidth}
    \centering
    \begin{tabular}{lc}
        \toprule
        \textbf{Algorithm}    & \textbf{PP} \\ \hline
          VDN (PS) & \CI{23.18}{-1.68,1.63} \\ \hline
          DVDN (GT) & \CB{23.20}{-1.00,1.40} \\
          IQL & \CI{23.00}{-1.79,1.72} \\ \bottomrule
    \end{tabular}
     \caption{Performance for predator-and-prey task, DVDN (GT) results on hyperparameter grid search trials.}\label{tab:replacement}
    \end{minipage}\hfill
\end{table}

\end{document}